\documentclass[preprint]{elsarticle}

\usepackage{amsmath}
\usepackage{amsfonts,amssymb}
\usepackage{bm}
\usepackage{cases}
\usepackage{subfigure}
\usepackage{booktabs}
\usepackage{multirow}
\usepackage{listings}
\usepackage{algorithm}
\usepackage{algpseudocode}
\usepackage{marvosym}
\usepackage{graphicx} %
\usepackage{epsfig}
\usepackage{fancyhdr}
\usepackage{setspace}
\usepackage{helvet}
\usepackage{makecell}

\makeatletter
\let\ps@pprintTitle\ps@plain
\makeatother

\begin{document}
\begin{frontmatter}
\title{Enhancing generalization in endwall film cooling prediction: Incorporating the superposition principle into transformer-based neural operators\tnoteref{published}}
\tnotetext[published]{Author manuscript. This article appeared in Physics of Fluids 36, 126110 (2024) and may be found at https://doi.org/10.1063/5.0239483.}

\author[1]{Qineng Wang}
\ead{zhet1997@stu.xjtu.edu.cn}
\author[1]{Liming Song}
\ead{songlm@xjtu.edu.cn}
\author[2,3]{Tianyuan Liu\corref{cor1}}
\ead{tianyuanliu1992@gmail.com}
\author[1]{Zhendong Guo\corref{cor2}}
\ead{guozhendong@xjtu.edu.cn}
\cortext[cor1]{Corresponding author}
\cortext[cor2]{Corresponding author}
\affiliation[1]{organization={Institute of Turbomachinery, Xi'an Jiaotong University},city={Xi'an, Shaanxi},postcode={710049},country={People's Republic of China}}
\affiliation[2]{organization={ENN Science and Technology Development China Co., Ltd.},city={Langfang, Hebei},postcode={065001},country={China}}
\affiliation[3]{organization={Hebei Key Laboratory of Compact Fusion},city={Langfang, Hebei},postcode={065001},country={China}}

\begin{abstract}

In this study, a physics-enhanced neural operator framework is proposed to enhance the generalization prediction ability of the cooling layout of a turbine endwall with variable number of film holes.
Specifically, inspired by the film cooling superposition principle, we propose a film cooling prediction model, namely superposition-based deep neural operator (SDNO), that divides the endwall temperature field prediction into two stages.
In the first stage, the cooling layout of a turbine endwall is divided into several sub-parts with randomly assigned film holes, and a Transformer-based neural operator network, namely Calculate Net, is designed to predict the temperature field of each sub-part.
Then, in the second stage, another neural operator network, i.e., Super Net, is trained to combine the temperature fields predicted by Calculate Net for each sub-part and obtain the superposed temperature field of the full cooling layout.
Additionally, instead of directly taking the film cooling contours as pixel plots, a signed distance function (SDF) which is sensitive to the variable locations of cooling holes, is designed to encode the location information of cooling holes.
Furthermore, the proposed endwall film cooling prediction model is trained with the samples that changing the number of film holes from 1-5 with variable locations.
Then, the trained prediction shows excellent generalization prediction ability, which can accurately predict the film effectiveness of the cooling layout with 10-20 film cooling holes that are unseen in the training samples.
The proposed SDNO also improves prediction accuracy relative to the fully supervised baseline.
With the above, the effectiveness of our proposed prediction model has been well demonstrated.

\end{abstract}

\begin{keyword}
Film cooling prediction; Physics-enhanced deep learning; Deep neural operator; superposition principle.
\end{keyword}

\end{frontmatter}

\section{Introduction}
\par
As the core equipment for the thermal and power conversion, a gas turbine plays a key role in transportation, the energy, and chemical industries, and other fields.
With the continuous improvement of the requirements for output power and thermal efficiency of gas turbines, the increase of Turbine Inlet Temperature (TIT) has become an important trend in technological development, which also results in turbine components bearing very serious high-temperature loads.
Researchers are required to develop more advanced thermal protection technologies to ensure the stable and efficient operation of gas turbines.
Among many cooling technologies, gas film cooling technology is widely used for thermal protection of turbine components. This technology avoids direct contact with metal parts at high temperatures by spraying cold gas through holes or gaps and forming a protective gas film, thus significantly reducing the heat load of the material.
\par
The efficiency of film cooling is affected by various factors, including the flow conditions of the main stream and coolant, the shape and layout of the film-cooling holes, and the geometry of the blades.
Especially in the end-wall area of the blade, the existence of complex three-dimensional flow characteristics such as horseshoe vortices, channel vortices, and corner vortices, along with the overall pressure gradient, greatly increases the difficulty of achieving effective gas film coverage.
Consequently, the prediction of the cooling effect of film-cooling holess in these complex flow environments becomes a difficult nonlinear problem.
Although computational fluid dynamics (CFD) offers the potential to solve such problems, it requires a lot of time and computational resources to deal with highly complex situations, prompting researchers to explore more efficient and less costly prediction methods.
\par
Researchers have gradually turned to more economical and efficient surrogate prediction methods to achieve rapid prediction of gas turbine end-wall film cooling effect.
Additionally, the fast prediction model, serving as an extremely low-cost alternative, can often be used for the optimization of film-cooling holess layout and other scenarios requiring large number of evaluations.
Traditional surrogates (such as GPR, RBF, SVM, etc.) are used to predict the global effect of end-wall air film cooling.
However, these models can only predict the average comprehensive cooling effect expressed as a scalar, which is difficult to accurately reflect the local temperature distribution on the blade surface, and there are certain limitations in capturing detailed performance when capturing local overheating.
\par
To overcome this limitation, fully supervised deep learning algorithms are introduced as a new research paradigm to enhance the accuracy and intuitiveness of the prediction of end-wall film cooling effects.
Within this framework, a series of innovative studies have been proposed.
For example, Li used the long short-term memory (LSTM) network to study the one-dimensional distribution of the effectiveness of flat plate gas film, which pioneered the first deep learning method for film cooling analysis\cite{yangEstablishmentLongshorttermmemoryModel2020a}.
Zhi used the PointNet to achieve two-dimensional distribution prediction of gas film effectiveness around a single film-cooling holes on a flat plate, taking into account the influence of boundary conditions\cite{zhiResearchPerformancePredictions2023}. This approach improved the applicability and prediction accuracy of the model under specific working conditions.
By building a convolutional neural network (CNN), Xing not only predicts the two-dimensional distribution around a single film-cooling holes but also establishes the reverse prediction models synchronously, demonstrating the strong potential of deep learning in complex film cooling analysis~\cite{xingDirectInverseModel2022}.
Wang further combined the interaction between the vortex generator and the film-cooling holess\cite{wangFastPredictionUncertainty2023}.
Moreover, they used the conditional generation adversarial network (cGAN) model to predict the overall temperature field of multiple film-cooling holess on the plate, providing a new solution for gas film cooling under complex geometric shapes and boundary conditions~\cite{wangTwodimensionalPredictionSuperposition2023}.
Other researchers, utilizing CNN\cite{liDeepLearningMethod2022}, Fourier neural operator (FNO)\cite{wangMetafourierNeuralOperators2024}, and UNet\cite{zhangInvestigationImplementationGenerating2024} respectively, further expanded the application of deep learning in the prediction of end-wall gas film cooling effect, especially in the prediction of gas film pore configuration and temperature field distribution, achieving remarkable results.
\par
Above results demonstrate that deep learning techniques possess significant potential for predicting the effect of film cooling in gas turbine end-wall regions, especially in complex flow environments and geometric configurations.
All of these studies demonstrate significant advantages in reducing computational cost and time, and maintain high prediction accuracy for in-distribution (ID) of the training data.
In current engineering design practices, there is an urgent need to enhance the generalization and universality of models while maintaining their accuracy.
However, including all possible designs in the data set is impractical when constructing a general-purpose film cooling prediction model, and purely data-driven models often experience a sharp decline in prediction accuracy when encountering samples for out-of-distribution (OOD) of the training data.
A concrete example of this issue is the prediction of the number of holes in gas film cooling; data-driven models struggle to predict distributions outside the training set unless they are augmented with physical mechanisms. For instance, studies collecting data from one to three rows of exhaust film holes and training the network accordingly demonstrate high prediction accuracy for targets comprising no more than three rows of film-cooling holess. However, when the model is applied to predict configurations exceeding three rows of gas film pores, the prediction error escalates to over 50\%. In contrast, traditional models based on physical principles exhibit robust adaptability to varying numbers of film-cooling holess and can accurately predict these numbers irrespective of the quantity involved. Due to significant variability in the number and layout of the discrete geometric features of film-cooling holes layouts, the prediction model suffers from substantial deficiencies in accuracy and extrapolation capabilities. Furthermore, building a model with generalization capabilities necessitates reliance on extensive data sets, often an unrealistic expectation in practical engineering applications.
\par
In view of this challenge, although fully supervised deep learning models have achieved notable results in predicting the efficiency of gas film cooling, their scope of application has been significantly limited due to their inadequate generalization ability and reliance on large-scale training data.
To overcome this bottleneck, this study proposes a comprehensive strategy that combines traditional physical theories, specifically, the film superposition principle, with advanced neural network techniques, particularly, neural operator networks.
By abstracting the film superposition principle and further representing it as an FNO network with undetermined parameters, the strategy of "decomposition-computing-superposition" is adopted for model training and inference. This method not only significantly improves the model's performance in terms of prediction accuracy and extrapolation ability but also provides an efficient and reliable solution for addressing highly complex temperature field prediction problems.
By integrating physical mechanisms and deep learning techniques, the method in this study demonstrates unique advantages in addressing complex situations beyond the capability of traditional models, providing new perspectives and tools for high-precision temperature field prediction in future engineering designs.
\par
This paper's primary contributions are as follows:
(1)
A method utilizing SDF function to address small-scale geometric features in flow field prediction is proposed for enhancing the prediction accuracy.
(2)
The superposition principle is integrated into the deep neural operator architecture, significantly enhancing the model's performance in terms of prediction accuracy and extrapolation ability for the gas film cooling prediction problem.
(3)
The training strategy of "decomposition-computation-superposition" tailored to the SDNO model, is formulated, offering an efficient and reliable solution for addressing the highly complex temperature field prediction problem.
\par
The remainder of the paper is organized as follows.
In Section \uppercase\expandafter{\romannumeral2}, the studied Pak-B blade end-wall air film cooling model is introduced in detail, including its numerical simulation model and parametric generation method, and the preparation method of the dataset is described.
Then, in the Section \uppercase\expandafter{\romannumeral3}, we employ the fully supervised model to complete the training of the prediction model and find that the fully data-driven method lacks the ability of extrapolation and generalization. It is also found that the Seller formula cannot be used to predict the temperature of the complex coupled heat transfer process by extrapolating the gas film effectiveness superposition principle.
In Section \uppercase\expandafter{\romannumeral4}, the network architecture of deep neural operators based on the principle of fusion superposition and its corresponding training and reasoning strategies are introduced in detail. Two training strategies, joint training and fine-tuning training, are introduced.
In Section \uppercase\expandafter{\romannumeral5}, we compare the deep neural operators with the basic model using different training mechanisms and prove the extrapolation ability of the superimposed model and the high efficiency of training.
Finally, the conclusion of this paper is summarized in Section \uppercase\expandafter{\romannumeral6}.

\section{Problem Setup}
\par
\begin{figure}[ht]
\begin{center}
\includegraphics[width=0.7\textwidth, trim = 0 0.5cm 0 0.5cm]{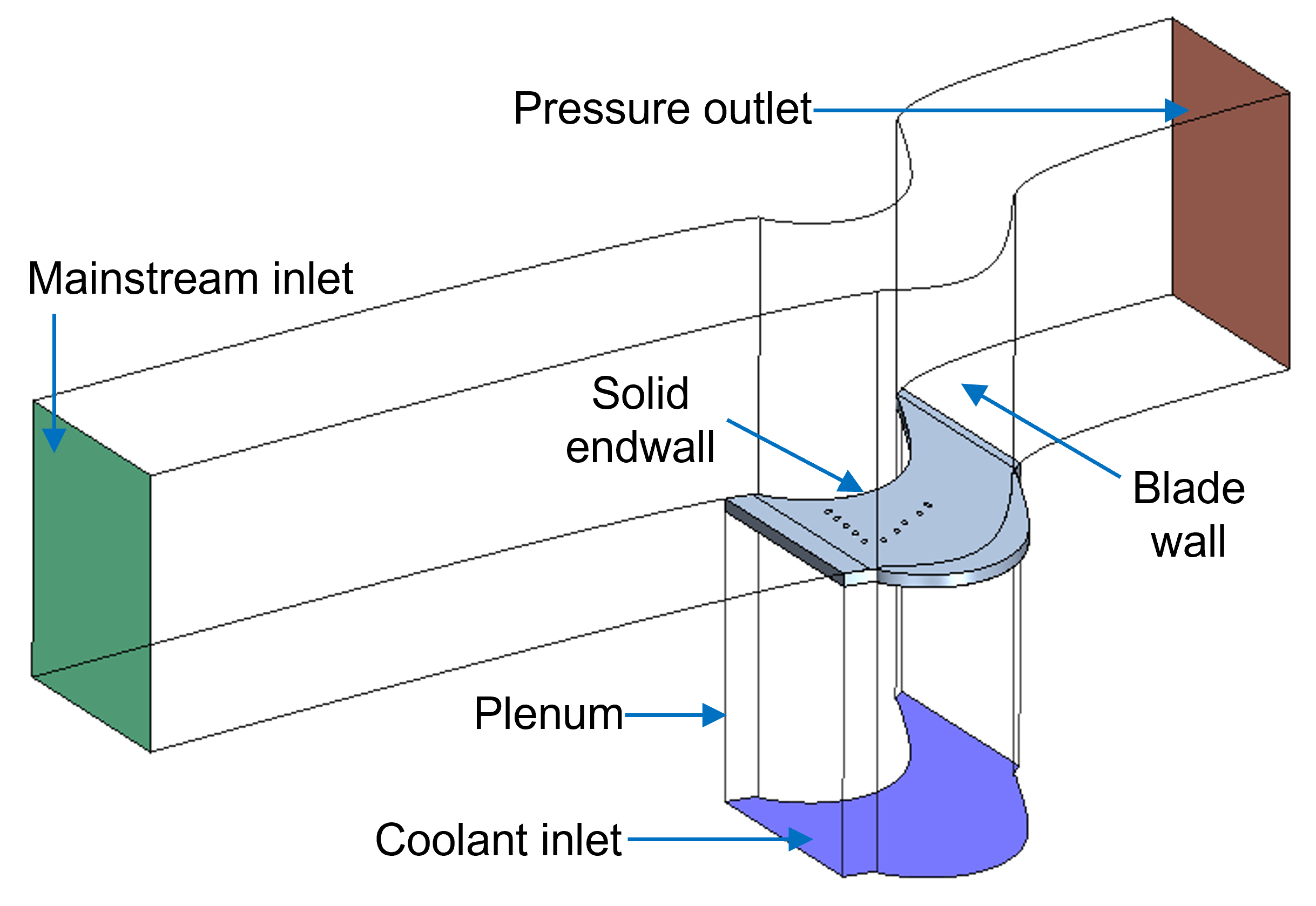}
\end{center}
\caption{The Pak-B blade endwall gas film cooling structure diagram}
\label{domain}
\end{figure}

This study focuses on the gas film-cooling holes layout on the endwall of the Pak-B blade.
In previous studies on end-wall gas film cooling, the Pak-B blade profile has been widely utilized as a research object due to its typical structure and flow characteristics.
The Pak-B blade components, flow passage, film-cooling holes on the end wall, and the plenum under the endwall are depicted in Fig. \ref{domain}.
In this article, the film-cooling holes layout of the Pak-B blade is evaluated using a high-fidelity flow-thermal coupling numerical simulation model.
To achieve robust generalization performance for the prediction model, a parametric generation technology was developed for film-cooling holes layout with extremely high degrees of freedom.
This was established while keeping the turbine blade profile and working conditions constant.
This technique is capable of generating arrangement schemes with varying numbers of film-cooling holess, thus providing a wide range of variable space for studying the gas film cooling effect.
Furthermore, building upon the above numerical model and parametric generation method, this study constructs a detailed dataset to investigate the effect of film-cooling holes layout on cooling efficiency via a data-driven approach.
In the following section, the construction of the numerical simulation model, the layout parametric generation method, and the preparation process of the dataset will be introduced in detail.
\subsection{Numerical simulation and validation}
\par
In this article, a computational model of coupled heat transfer is used to obtain the temperature field distribution under different gas film-cooling layout schemes.
In the calculation domain depicted in Figure \ref{domain}, the high-temperature fluid enters through the main inlet, while the coolant enters through the inlet of the plenum and is ejected through the holes on the endwall to mix with the main flow.
The cold air reduces the endwall temperature by forming a gas film covering on one hand, and by facilitating heat transfer on the other hand.
So, the entire process is more complicated than the adiabatic model.
The specific settings of the above numerical simulation model are shown in Table \ref{designCondition}.
The Star-CCM+ is used for generation of a prismatic mesh and the calculation of the finite volume method.
The number of grid nodes for the different samples ranges between 2.8 million and 3.6 million.
The computational time required to complete a sample using 4 cores in parallel is approximately 3 hours.
More specific details of the numerical model can be seen in our previous paper~\cite{wangNOVELMULTIFIDELITYSURROGATE2023b}.

\begin{table}[!ht]
  \centering
  \caption{Design conditions of the Pak-B blade}
  \begin{tabular}{lcc}
  \hline
      Boundary Condition & Symbol & Value \\ \hline
      Mainstream outlet Reynolds number & $Re_\text{out}$ & 1.98 $\times 10^5$ \\
      Mainstream inlet Mach number & $Ma_\text{in}$ & 0.029 \\
      Mainstream outlet Mach number & $Ma_\text{out}$ & 0.047 \\
      Mainstream inlet total temperature & $T_{\infty}$ & 323K \\
      Mainstream inlet turbulence intensity & $T_u$ & 6\% \\
      Velocity boundary layer thickness & $\delta$ & 33mm \\
      Momentum boundary layer thickness & $\theta$ & 2.5mm \\
      Inlet total temperature of cooling gas & $T_c$ & 286K \\ \hline
  \end{tabular}
  \label{designCondition}%
\end{table}

\subsection{Film-cooling holes layout generation}
\par
In order to construct a comprehensive and efficient data set, a film-cooling holes layout parameterization technique with high flexibility was developed in this study.
Based on a series of normalized spatial relative position parameters, this technique can generate a variety of reasonable film-cooling holes layout schemes.
\begin{figure}[ht]
\begin{center}
\includegraphics[width=0.65\textwidth, trim = 0 0.5cm 0 0.5cm]{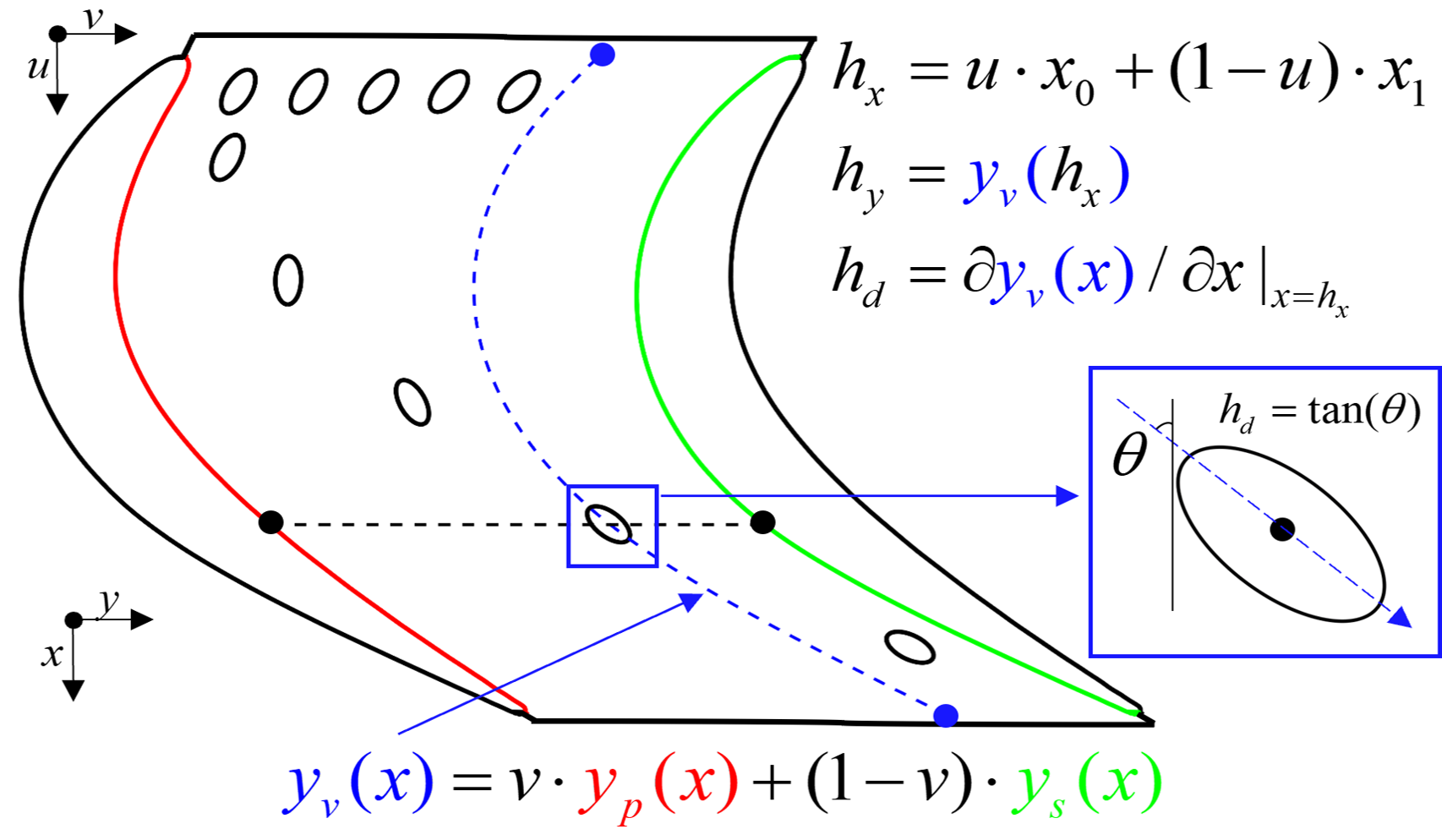}
\end{center}
\caption{Layout mode generates schematic diagram}
\label{layout}
\end{figure}
The development of this parameterization method accounts for the variability and complexity inherent in the layout of film-cooling holes in practical engineering design.
Its objective is to establish a robust data foundation for predicting the cooling efficiency of turbine blades.
The method comprises three essential steps:
\par
(1)
Holes distribution generation:
This step entails generating the relative position coordinates of the film-cooling holes, ensuring that each hole's position can be defined using a standardized set of parameters.
Meanwhile, the generation process also limits the relative distance between the two holes to be greater than 0.15, to avoid mutual interference during geometric modeling.
\begin{equation}
\begin{array}{l}
  ({u_i},{v_i}) \in \{ U,V\} \quad (i = 1, \cdots ,{n_H})\\
  \left\{ {\begin{array}{*{20}{c}}
  {\forall i \ne j,\quad {{({u_i} - {u_j})}^2} + {{({v_i} - {v_j})}^2} < {{0.15}^2}}\\
  {{u_i} \in [0.05,0.8]}\\
  {{v_i} \in [0.05,0.95]}
  \end{array}} \right.
  \end{array}
\end{equation}
\par
(2)
Physical coordinate mapping:
In this step, the relative position coordinates acquired in the preceding stage are mapped to physical space and transformed into absolute coordinates on the turbine's endwall.
\begin{equation}
    \begin{array}{l}
  {f_v}( \cdot ) = v \cdot {f_p}( \cdot ) + (1 - v) \cdot {f_s}( \cdot )\\
  \left\{ {\begin{array}{*{20}{c}}
  {x = u \cdot ({x_{{\rm{max}}}} - {x_{{\rm{min}}}}) + {x_{{\rm{min}}}}}\\
  {y = {f_v}(x)}
  \end{array}} \right.
  \end{array}
\end{equation}
\par
(3)
Rotation angle definition: Based on the absolute coordinates of the hole on the blade's endwall, this step determines the injection angle of each hole, thereby refining the geometric characteristics of the film-cooling holes.
\begin{equation}
  \theta  = {\tan ^{ - 1}}(\partial {f_v}(x)/\partial x)
  \end{equation}
\par
Through the described methodology, the algorithm automatically calculates the center coordinates and circumferential deflection angles of all holes based on the input of a specific number, denoted as $h_k = {x_k, y_k, \theta_k}$.
For model simplification and consistency, the diameter of each hole was set at 4 mm and the inclination angle at 30 degrees.
This configuration ensures that all geometric parameters of the film-cooling holes are fully defined, thereby preventing potential interactions between holes.
The parametric method proposed in this study comprehensively addresses the requirements of practical engineering applications, while also encompassing a broad spectrum of design spaces through the implementation of highly flexible parameter control mechanisms.
\subsection{Data preparation}
\par
The focal point of this study is to assess the extrapolation and generalization capabilities of deep learning models integrating physical mechanisms, particularly regarding the increased number of film-cooling holes.
"Extrapolation generalization" denotes a model's capacity to adapt to novel scenarios beyond the confines of its training dataset.
Accordingly, in constructing the dataset, a deliberate choice was made to compose a training set comprising samples with a relatively low number of film-cooling holes, whereas the validation set was meticulously curated to feature a significantly higher number of film-cooling holes than those encompassed within the training set's range.
To precisely investigate and quantify the model's extrapolation capability, this study employed a non-mixed strategy in dataset composition.
Specifically, samples featuring varying numbers of film-cooling holess in the training and validation sets were strictly segregated without overlap.
This approach aims to establish a distinct boundary for assessing the model's capacity to handle previously unobserved data.
\begin{figure}[ht]
\begin{center}
\includegraphics[width=0.95\textwidth, trim = 0 0 0 0]{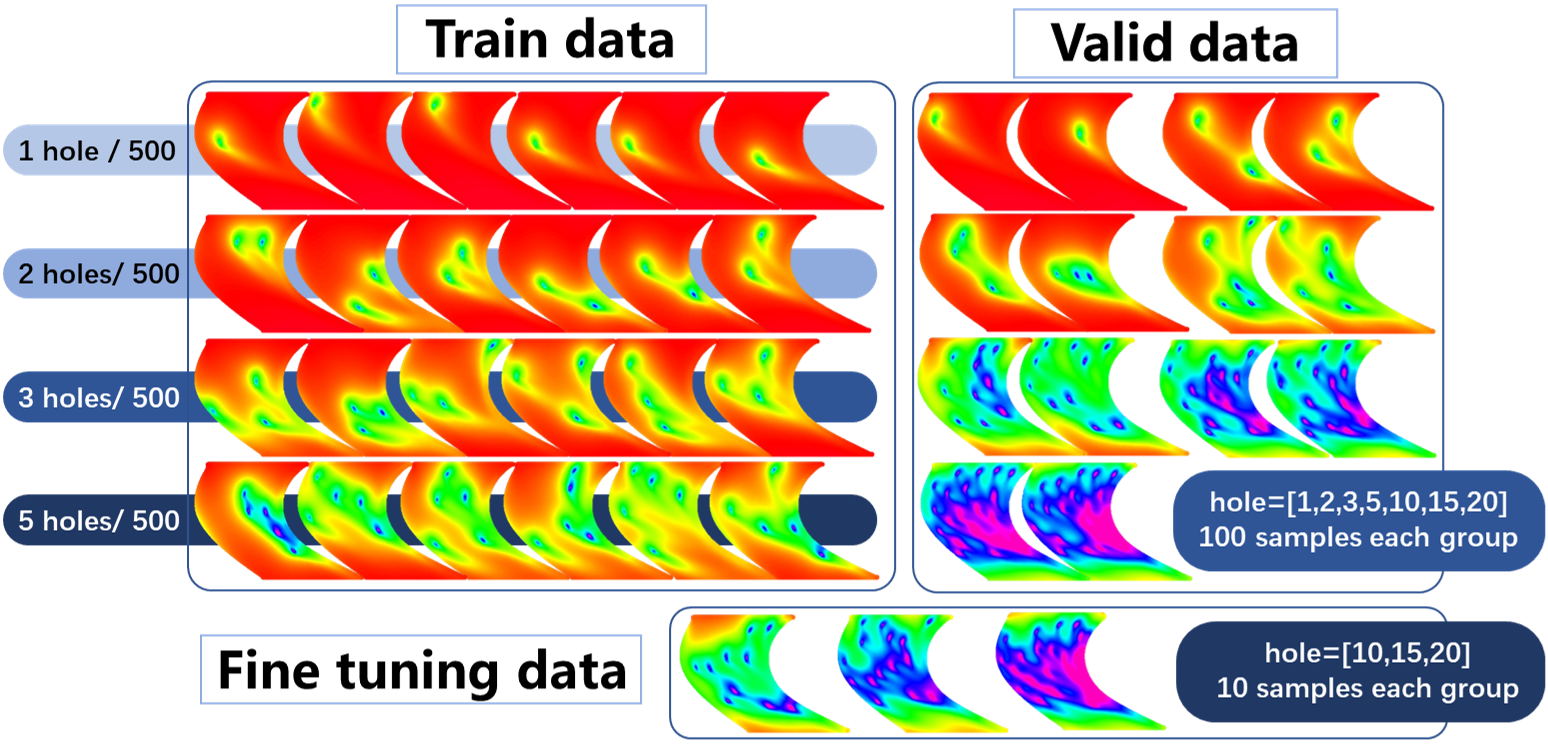}
\end{center}
\caption{Non-mixed data set preparation strategy considering the number of film-cooling holes on endwall}
\label{dataset}
\end{figure}
\par
Employing the parameterization technique outlined in Section 2.2 of this paper, a series of geometric models with varying holes numbers on the endwall are generated.
Specifically, 600 sets of data samples were generated for configurations with film-cooling holes numbers of 1, 2, 3, and 5, respectively. Additionally, for scenarios with more of film-cooling holes, namely configurations featuring 10, 15, and 20 holes, 110 sets of samples were generated for each configuration.
Subsequently, these data samples underwent processing using the numerical calculation method outlined and validated in Section 2.1 of this paper to assess the temperature field distribution on the turbine endwall surface.
In general, as illustrated in Fig.\ref{dataset}, the 2730 sets of prepared data were partitioned into three subsets: a training set comprising 2000 samples, encompassing samples with film-cooling holes numbers of 1, 2, 3, and 5, with 500 samples allocated for each configuration;
a validation set comprising 700 samples, spanning samples with all film-cooling holes numbers configurations, with 100 samples allocated for each configuration;
and finally, a fine-tuning test set comprising 30 samples tailored specifically for scenarios with a higher holes numbers.
Through above dataset design, this study aims to further explore and validate the prediction model's potential application in enhancing the efficiency of turbine blade end-wall air film cooling.
Specifically, the study seeks to assess the accuracy and reliability of the model in handling complex scenarios beyond the scope of its training data.
\section{Fully Supervised training and Empirical Superposition Principle}
\par
Before examining the deep neural operator architecture of the proposed SDNO network, the prediction effect of the fully supervised learning network and the superposition principle of film-cooling holes on the studied objects are first thoroughly investigated.
In this section, a Transformer network $\hat{\mathcal{C}}(\dot)$ is constructed and subsequently trained using the training set data delineated in Section 2.3.
Simultaneously, the interpolation accuracy and extrapolation generalization of the model are assessed utilizing the validation set data. Subsequently, the extrapolation prediction of the number of film-cooling holes is directly tested using the traditional superposition principle to evaluate its practical effect on the Pak-B blade's endwall.

\subsection{Data preprocessing}
\par
\begin{figure}[ht]
\begin{center}
\includegraphics[width=0.9\textwidth, trim = 0 0 0 0]{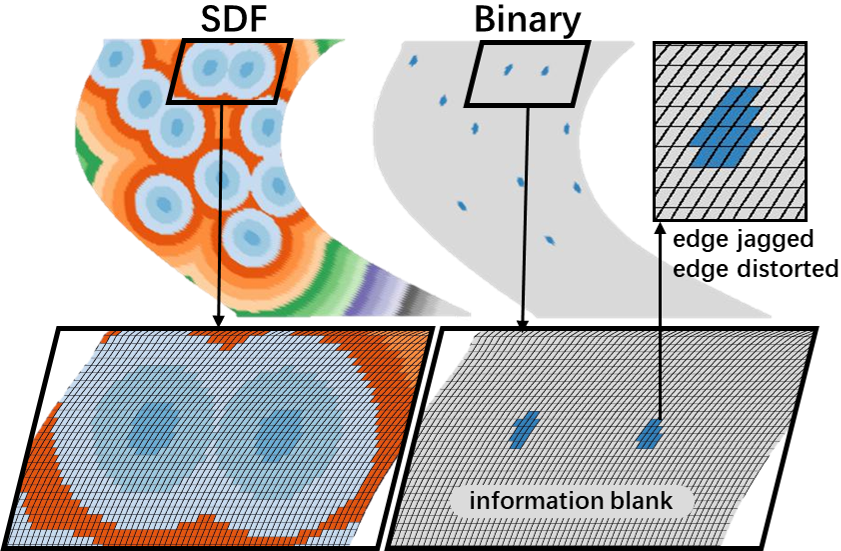}
\end{center}
\caption{The comparison of traditional binary method and SDF method in preprocessing}
\label{sdf_method}
\end{figure}

The data of film-cooling holes layout on the end wall is performed with preprocessing, before input into the neural network.
Traditionally, the film-cooling holes' layout is processed using a binary method when it is transformed into discrete data points, using 0 and 1 to denote the inside and outside location node of the holes, respectively.
This approach can lead to boundary jaggedness and information voids, when dealing with small-scale features.
For instance, as depicted on the right in Fig. \ref{sdf_method}, an elliptical edge of hole, due to its small size, might be represented by only 10 or more pixels, with many nearby pixels failing to convey effective information, resulting in a jagged boundary.
\par
To address above issues, the Signed Distance Function (SDF) method is employed to preprocess the geometry information, mitigating the problems of boundary jaggedness and information voids.
The detailed implementation of the SDF method for elliptical boundaries is documented in reference~\cite{eberlyDistancePointEllipse}.
As illustrated on the left side of Fig. \ref{sdf_method}, using SDF processing, the value recorded by each pixel represents the distance to the elliptical edge of the nearest film-cooling hole.
This technique effectively resolves the issues of boundary jaggedness and information gaps, enhancing the efficiency with which gas film hole configuration information is input into the deep neural operator network.
\par
In this article, a fixed struct $128\times128$ grid $G = {g_{ij}} (i,j\in[1,128])$ is set on the endwall surface. From each single film cooling hole, the SDF value can be calculated as $x_{k,ij} = \text{sdf}(h_k, g_{ij})$, and the entire SDF field can also be calculated with all sub SDF fields as $x_{ij}=\min(x_{1,ij}, \cdots, x_{n_H,ij})$.
\subsection{Establishment and Training of Fully Supervised Network}
\label{sec:baseline}
The Transformer-based neural operator $\hat{\mathcal{C}}$ predicts the endwall temperature field from the film-cooling layout.
Its input combines one SDF channel and two spatial-coordinate channels on the $128\times128$ grid, giving an input tensor of size $N\times128\times128\times3$, where $N$ is the batch dimension.
As shown in Table~\ref{tab:transformer}, the spatial grid is represented by 16384 tokens with 64 features per token. The network uses three Galerkin-based simplified linear-attention layers, with residual connections and multilayer perceptrons; the simplified attention avoids the softmax computation.
The final layers produce one temperature value at each grid point.
To examine extrapolation, the training set is progressively expanded from single-hole samples to samples with 1, 2, 3, and 5 holes.
\begin{table}[htbp]
\centering
\caption{The detailed network structure of Transformer.}
\label{tab:transformer}
\resizebox{\textwidth}{!}{%
\begin{tabular}{ll}
\toprule
Network type & Output size \\ \midrule
Input & $N \times 128 \times 128 \times 3$ \\
(Linear + RELU) $\times$ 5 & $N \times 16384 \times 64$ \\
Linear attention + residual & $N \times 16384 \times 64$ \\
LayerNorm + linear + RELU & $N \times 16384 \times 64$ \\
Linear attention + residual & $N \times 16384 \times 64$ \\
LayerNorm + linear + RELU & $N \times 16384 \times 64$ \\
Linear-Attention + residual & $N \times 16384 \times 64$ \\
LayerNorm + linear + RELU & $N \times 16384 \times 64$ \\
(Linear + RELU) $\times$ 4 & $N \times 128 \times 128 \times 1$ \\
\midrule
\multicolumn{2}{c}{Total parameters: 103361} \\
\multicolumn{2}{c}{Parameter size: 0.39 MB; forward pass size: 180.34 MB} \\
\bottomrule
\end{tabular}}
\end{table}

Prediction accuracy is evaluated using the field average absolute error,
\begin{equation}
 e_{\mathrm{FAE}}=\frac{1}{N_{\mathrm{valid}}M}
 \sum_{i=1}^{N_{\mathrm{valid}}}\sum_{j=1}^{M}
 \left|f_{b,ij}-\widehat f_{b,ij}\right|,
 \label{eq:fae}
\end{equation}
where $f_{b,ij}$ and $\widehat f_{b,ij}$ are the reference and predicted temperature at grid point $j$ of validation sample $i$, and $M$ is the number of evaluated spatial points. Thus $e_{\mathrm{FAE}}$ is expressed in kelvin. The heat maps display $\log_{10}e_{\mathrm{FAE}}$.
\par
\begin{figure}[ht]
\begin{center}
\includegraphics[width=0.7\textwidth, trim = 0 0 0 0]{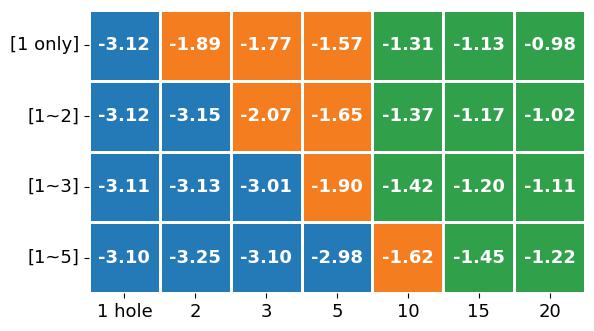}
\end{center}
\caption{The prediction error for the fully supervised baseline network}
\label{heat_base}
\end{figure}

Figure \ref{heat_base} displays the test accuracy of the validation set with a varying number of air film holes as the training set. Notably, the step-like distribution depicted in the figure clearly illustrates the model's high-precision prediction capability within the training set. However, within this data-driven, fully supervised learning framework, the model's extrapolation and generalization capabilities are limited. When the number of holes in the prediction object exceeds the range of the training set, the error increases dramatically, by 20 to 100 times. This phenomenon aligns with the findings of many related studies. Additionally, it is noteworthy that, even with the same sample size, an increase in the number of film-cooling holes results in a significant error increase, indicating that the model struggles to predict more complex gas film configurations.
\subsection{Film cooling superposition Principle}
Seller initially proposed the superposition principle of gas film-cooling effectiveness in 1965~\cite{sellersGASEOUSFILMCOOLING1963}. Subsequently, many researchers have modified and improved it~\cite{sasakiFilmCoolingEffectiveness1979, chen2DPredictionMethod2021}, the specific formula is presented below:
\begin{equation}
\begin{array}{l}
  \eta (\omega ) = \frac{{{T_\infty } - T(\omega )}}{{{T_\infty } - {T_c}}}\\
  1 - {\eta _{AB}}(\omega ) = (1 - {\eta _A}(\omega )) \cdot (1 - {\eta _B}(\omega ))
  \end{array}
\end{equation}
Here, $\omega$ refers to any spatial point within the computation domain, where $\omega \in \Omega$;
This principle enables to predict the temperature field of multiple holes by superimposing the temperature fields of individual holes, thereby evaluating the effects of multiple film-cooling holes, which significantly reduces computational costs.
However, subsequent research has indicated that this principle accurately predicts the superposition effects only in simpler scenarios~\cite{andreiniFilmCoolingSystem2005,kirollosEnergyBasedMethodPredicting2015,yangMachineLearningApproach2021}, with its applicability being limited in complex flow and heat transfer situations.
\par
\begin{figure}[ht]
\begin{center}
\includegraphics[width=0.9\textwidth, trim = 0 0 0 0]{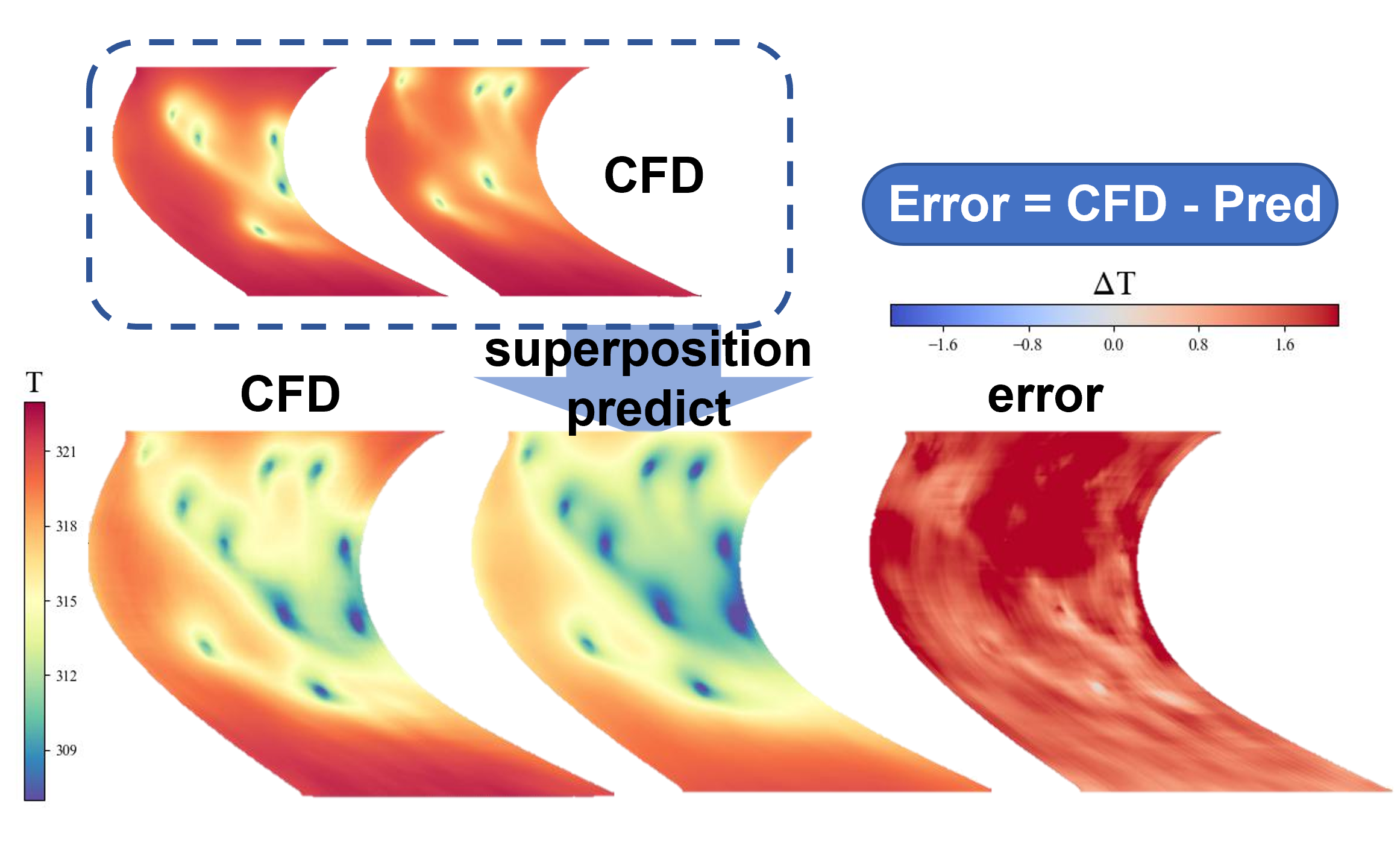}
\end{center}
\caption{Using Seller's formula in the extrapolation prediction of the film-cooling at the endwall of the pak-B blade.}
\label{seller}
\end{figure}

Upon direct application of the Seller superposition principle to the subject of this study, it was found that the principle struggles to accommodate complex flow fields and heat transfer dynamics.
Figure \ref{seller} illustrates the results of the Seller formula for extrapolating the prediction of the air film hole at the end wall of the Pak-B blade.
As depicted in the figure, the flow field temperature predicted by Seller's formula is notably lower than that calculated by CFD simulation.
This disparity arises because the reduction in the endwall surface temperature is attributed to both gas film coverage and heat transfer. However, the traditional Seller formula attributes all cooling solely to gas film coverage, resulting in an overestimation of the gas film's cooling capacity and an underestimation of the superposed temperature.
\par
In summary, the fully supervised model and the gas film superposition principle exhibit limitations in the context of complex flow environments and heat transfer considerations.
While fully supervised deep learning models can achieve high-precision predictions, they lack the capability to extrapolate and generalize.
However, the traditional validity of the superposition principle fails to meet the extrapolation needs for complex objects.
Therefore, to meet the objectives of this paper, integrating physical mechanisms into deep learning architectures is essential.
\section{Deep Operator Architecture Integrating Superposition Principle}
\subsection{Generalization and Concretization of the Film cooling Superposition Principle}
\par
Due to the weak extrapolation ability of the prediction model discussed in Section~\ref{sec:baseline},  a novel superposition-based deep learning prediction structure is proposed.
The following section will provide a brief introduction about how to applicator the superposition principle in the deep learning of gas film cooling prediction model.
The original Seller's formula can be expressed as a binary calculation, as follows:
\begin{equation}
  \begin{array}{l}
    \eta_C(\omega)= \frac{T(\omega)-T_{c}}{T_{\infty}-T_{c}}\\
    T^{(AB)}(\omega) =\eta_C^{-1} [\eta_{C}(T^{(A)}(\omega) )\cdot \eta_{C}(T^{(B)}(\omega)) ]\\
      = \text{Seller}\langle T^{(A)}(\omega), T^{(B)}(\omega) \rangle\\
    \end{array}
  \end{equation}

Recognizing that the superposition principle may not be applicable in most complex conditions, the overstrong hypothesis is removed and generalized.
(1) The assumption of explicit direct computation is removed, generally reducing it to a binary computation operation conforming to the commutative and associative laws;
(2) The assumption of computation localization is removed, generalizing the operand from individual space points $\omega$ to a certain scale space region $\Omega_p$.
After generalization, the formula for the superposition principle is as follows:
\begin{equation}
    T^{(AB)}(\Omega_p) = \text{Super}\langle T^{(A)}(\Omega_p) , T^{(B)}(\Omega_p) ; \theta_S \rangle \\
\end{equation}
Here $\Omega_p$ refers to continuous subfields in the computing domain, there are $\Omega_p \subset \Omega$;
Further, the general superposition principle can be concretized as an FNO network structure $\hat{\mathcal{S}}$, which is denoted as superposition network.
The specific network structure applicable in this article is shown in Table \ref{FNO}.
The input and output of the superposition network are normalized physical fields. Its input tensor has size $N\times128\times128\times2$, representing two fields to be combined. Patch partitioning produces $16N\times32\times32\times2$ tensors, as in a windowed Fourier transform, to capture local interactions at reduced computational cost. After two spectral-convolution blocks, the patch-reverse operation reconstructs a single temperature field of size $N\times128\times128\times1$.
The trainable superposition operator should retain the commutative and associative properties of the original principle. Input-channel order can be shuffled, or the network can be trained through iterative superposition, to encourage these properties.
\begin{equation}
  T^{(AB)}(\Omega_p) =\hat{\mathcal{S}}(T^{(A)},T^{(B)}; \theta_S)(\Omega_p)\\
\end{equation}
\begin{table}[htbp]
\centering
\caption{The detailed network structure of FNO.}
\label{FNO}
\resizebox{\textwidth}{!}{%
\begin{tabular}{ll}
\toprule
Network type & Output size \\ \midrule
Input & $N \times 128 \times 128 \times 2$ \\
Patch & $16N \times 32 \times 32 \times 2$ \\
Linear-1 & $16N \times 32 \times 32 \times 64$ \\
(Conv2d + GELU + SpectralConv2d) $\times$ 2 & $16N \times 64 \times 32 \times 32$ \\
Linear-2 & $16N \times 128 \times 32 \times 32$ \\
Linear-3 & $16N \times 32 \times 32 \times 1$ \\
Patch-reverse & $N \times 128 \times 128 \times 1$ \\
\midrule
\multicolumn{2}{c}{Total parameters: 17089} \\
\multicolumn{2}{c}{Parameter size: 0.07 MB; forward pass size: 0.3 MB} \\
\bottomrule
\end{tabular}}
\end{table}
\par
For the entire predictive deep learning network, the Transformer-based calculation network $\hat{\mathcal{C}}(\dot)$ introduced in Section~\ref{sec:baseline} continues to be employed for temperature field prediction.
The calculation network $\hat{\mathcal{C}}(\dot)$ and superposition network $\hat{\mathcal{S}}(\dot)$ together form the architecture of the SDNO. The expression of the proposed network is as follows:
\begin{equation}
  \hat{\mathcal{F}}\circ Sdf^{(AB)}(\Omega_p)
  =\hat{\mathcal{S}}(\hat{\mathcal{C}}\circ Sdf^{(A)}, \hat{\mathcal{C}}\circ Sdf^{(B)}) (\Omega_p)
\end{equation}
\subsection{The training strategy of superposition neural network}
\label{sec:training}
\par
The central focus of this research is to develop a comprehensive network architecture for predicting the temperature field distribution during the film cooling process using deep learning technology.
This architecture comprises two essential components: a computational network $\hat{\mathcal{C}}(\dot)$ that calculates the temperature field using input gas film pore configuration information, and a superposition network $\hat{\mathcal{S}}(\dot)$ that combines physical field results based on the superposition principle.
The integration of these two networks aims to address the intricate challenge of gas film cooling, particularly in predicting temperature field within the OOD of training dataset.
\begin{figure}[ht]
\begin{center}
\includegraphics[width=0.95\textwidth, trim = 0 0 0 0]{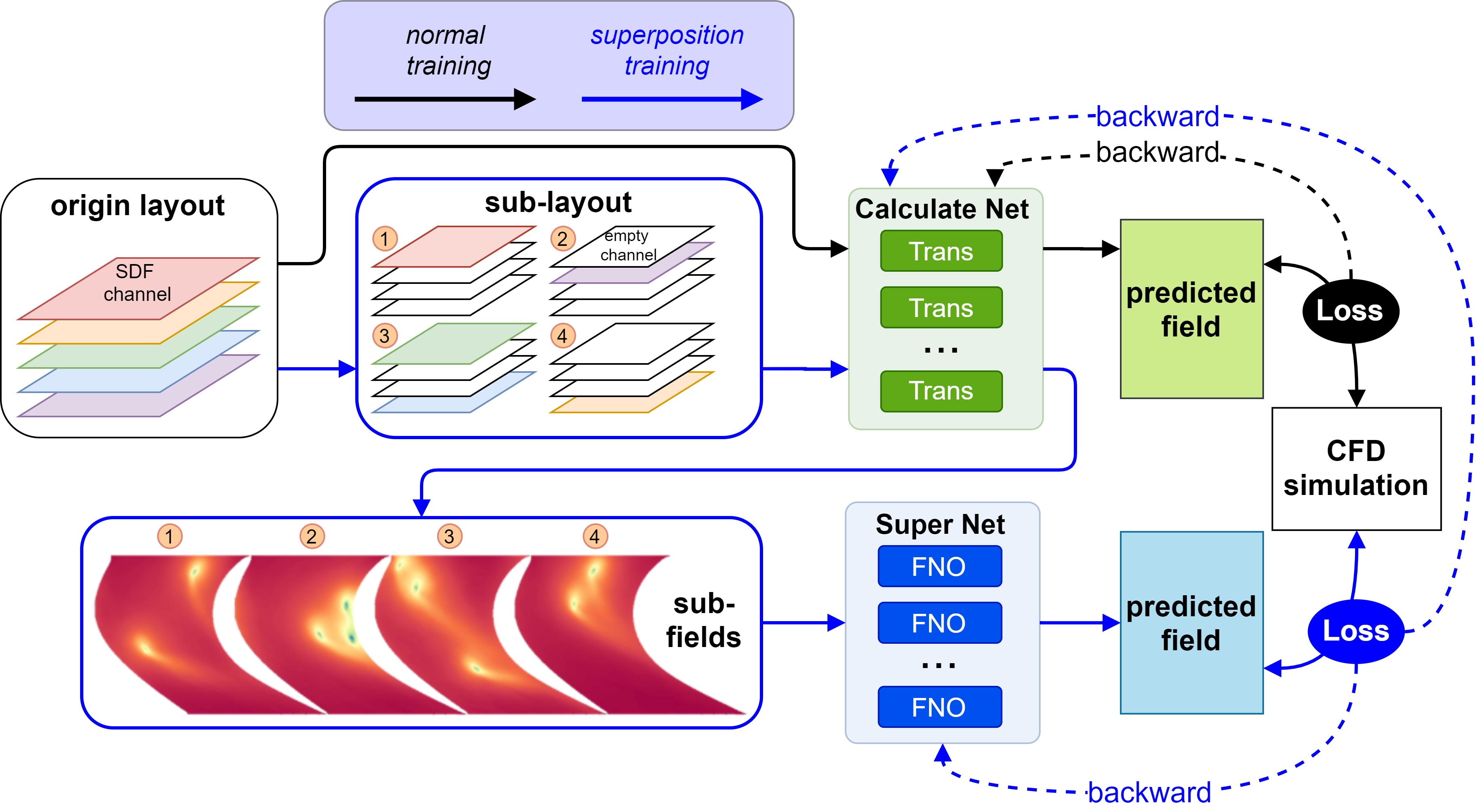}
\end{center}
\caption{The SDNO network structure and training strategy}
\label{net_structure}
\end{figure}
As depicted in Fig. \ref{net_structure}, the SDNO networks describes two distinct processing methods for the forward computation of the same sample. %
The standard mode (illustrated by the black arrow path) exclusively utilizes computational networks, aligning with the fully supervised approach outlined in Section 3.2.
While in the superposition mode (illustrated by the blue arrow path), the entire process is delineated into three essential steps of "decomposition - calculation - superposition" for detailed elucidation:
\par
(1)
Decomposition step:
During this stage, the layout scheme of film-cooling holes within the original sample is meticulously divided into $N$ subsamples, ensuring that the total number of gas film pores $(n_H)$ in the original sample is fully represented without duplication. This decentralized process is devised to enhance computational efficiency, where $N=2^{t_{\text{super}}}$ to accommodate the requirements of the superposition network.
Here, $t_{\text{super}}$ is denoted as superposition times.
\par
(2)
Calculation step:
During this phase, all decomposed sub-sample features are fed into the computational network $\hat{\mathcal{C}}(\dot)$, which subsequently produces the corresponding $N$ predicted temperature field results to provide essential data support for the superposition step.
\par
(3)
Superposition step:
Following the computation of the sub-sample temperature fields, the superposition network combines them through $t_{\text{super}}$ levels of pairwise superposition to obtain one final temperature field.
\par
Additionally, upon completion of the forward inference process, the parameters of both the computational network $\Theta_{\mathcal{C}}$ and the superposition network $\Theta_{\mathcal{S}}$ are updated through the backward process (as depicted by the blue dashed arrow in Fig. \ref{net_structure}).
This dual network parameter updating mechanism ensures that, even when utilizing the same training data, the training outcomes of the computational network will differ from the fully supervised training results in Section 3.2 due to the influence of the superposition network.
This approach enhances both the accuracy of prediction and the generalization capability of the model for complex gas film cooling by simulating the superposition effect of real physical processes.
\begin{figure}[htbp]
\centering
\subfigure[$t_{\text{super}}=1, \, N_{\text{sub}}=2$]{
\begin{minipage}[t]{0.7\linewidth}
\centering
\includegraphics[width=1\textwidth]{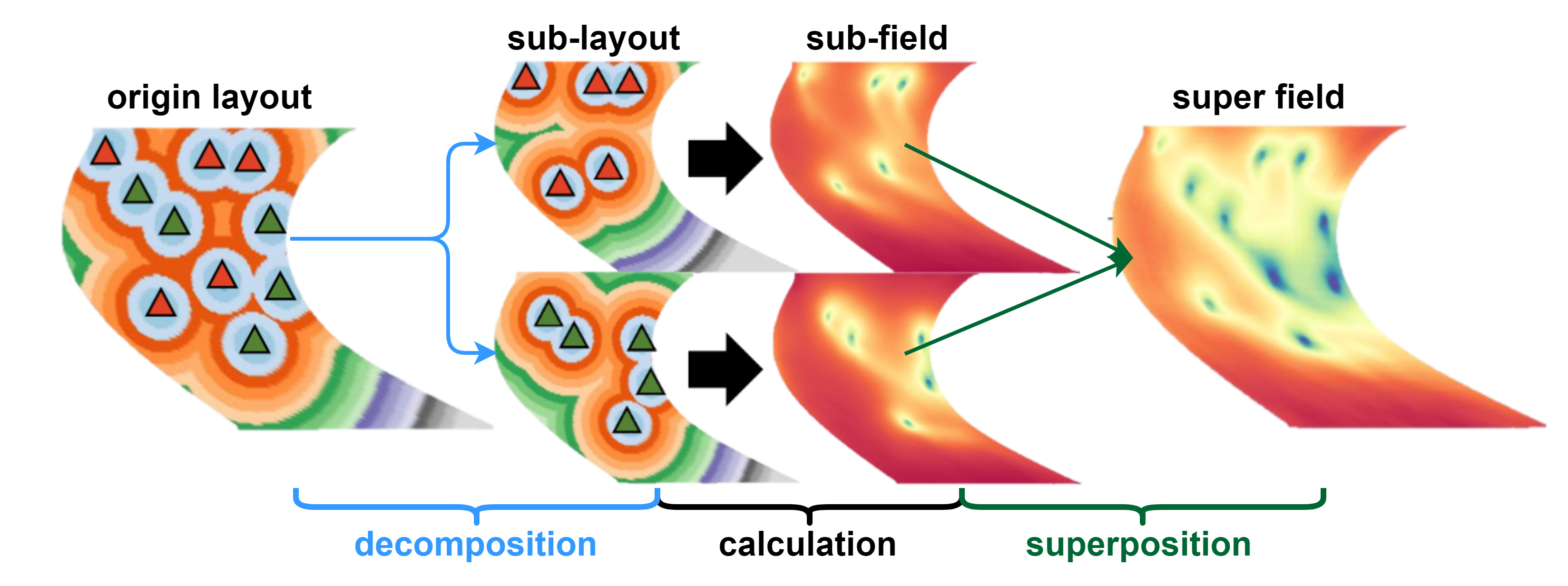}
\end{minipage}%
}%

\subfigure[$t_{\text{super}}=2, \, N_{\text{sub}}=4$]{
\begin{minipage}[t]{0.7\linewidth}
\centering
\includegraphics[width=1\textwidth]{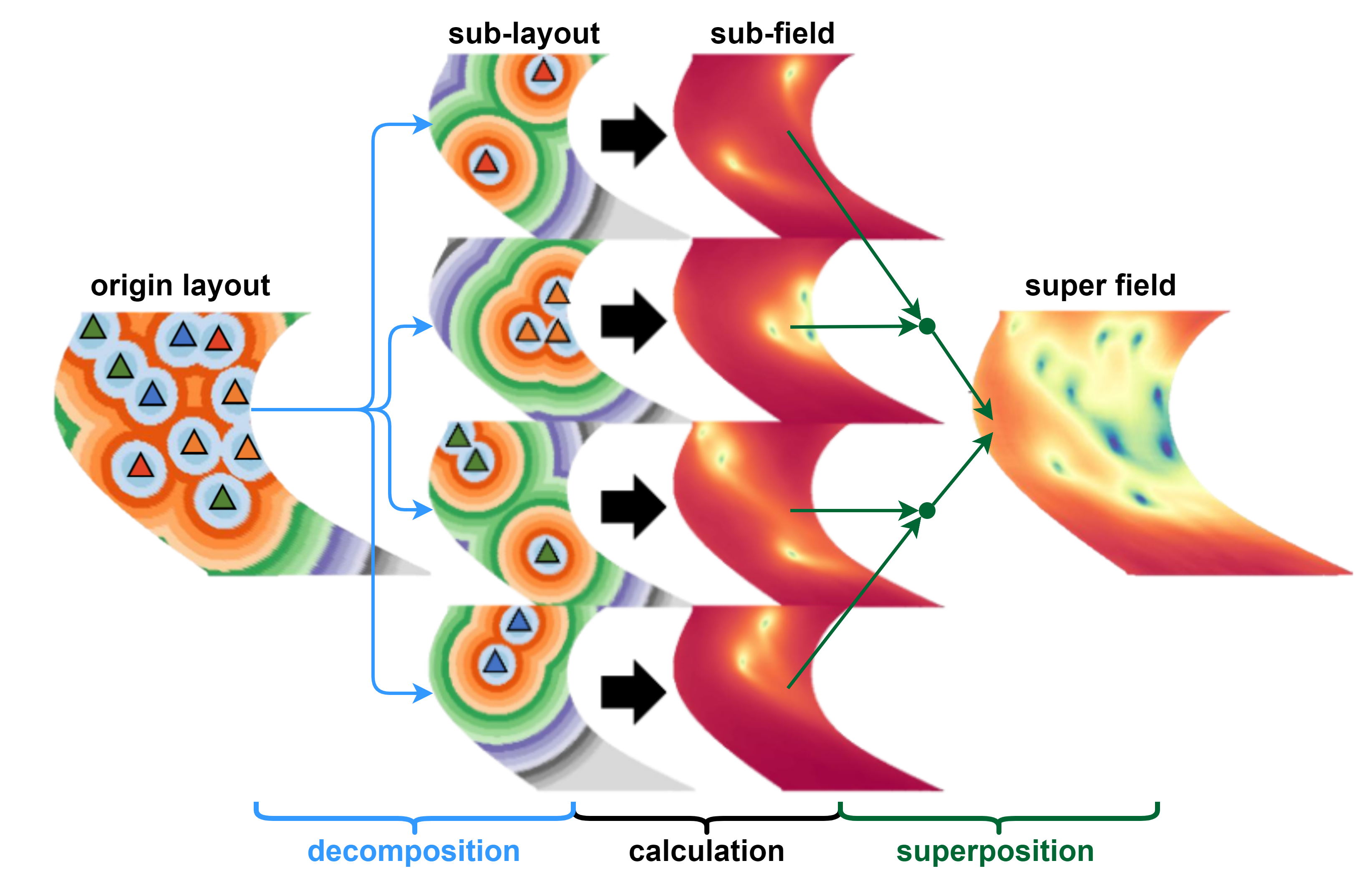}
\end{minipage}
}%
\centering
\caption{The sketch map for the decomposition-calculation-superposition steps for a 10-holes sample}
\label{super_step}
\end{figure}
The details of a superposition prediction process are illustrated in Fig. \ref{super_step}.
In the case where the superposition time $t_{\text{super}}$ is set to 1, as depicted in Fig. \ref{super_step}(a), the original sample containing 10 film-cooling holes is decomposed and transformed into two independent sub-samples.
These sub-samples are subsequently input into the computational network $\hat{\mathcal{C}}(\dot)$ to derive the predicted temperature field for each configuration.
Afterwards, the resulting two temperature fields are combined in the superposition network $\hat{\mathcal{S}}(\dot)$ to generate a unified temperature field.
Additionally, Fig. \ref{super_step}(b) illustrates a scenario where the $t_{\text{super}}$ increases to 2.
In this configuration, the original sample is divided into four sub-samples and undergoes two superimpositions to generate the final temperature field prediction.
Remarkably, with adequate hardware resources, the potential for increasing the $t_{\text{super}}$ further arises, thereby enabling precise predictions for increasingly complex samples.
\par
From Figure \ref{super_step}, it is evident that for the same sample, there exist multiple potential prediction paths by altering the decomposition method and superposition strategy.
In these prediction methodologies, the same sample contributes to the training of model parameters in various capacities, and this characteristic can significantly enhance model generalization.
Furthermore, it diminishes the necessary sample requirements, a particularly crucial aspect in domains where data acquisition is costly and limited.
\begin{equation}
\begin{aligned}
 \Theta_{\mathcal C}^{*}
 &=\underset{\Theta_{\mathcal C}}{\arg\min}\,
 \mathbb E_{\mathcal D}[\mathcal L_{\mathrm{normal}}],
 &\mathcal L_{\mathrm{normal}}&=\|\hat{\mathcal C}(X)-Y\|_2,\\
 (\Theta_{\mathcal C}^{*},\Theta_{\mathcal S}^{*})
 &=\underset{\Theta_{\mathcal C},\Theta_{\mathcal S}}{\arg\min}\,
 \mathbb E_{\mathcal D}[\mathcal L_{\mathrm{super}}],\\
 \mathcal L_{\mathrm{super}}
 &=\|\hat{\mathcal S}(\hat{\mathcal C}(X^{(A)}),\hat{\mathcal C}(X^{(B)}))-Y\|_2.
\end{aligned}
\label{eq:training_losses}
\end{equation}
The normal and superposition paths optimize the two losses in Eq.~\ref{eq:training_losses}. Table~\ref{tab:training_strategy} specifies the forward operation, updated parameters, learning rate, and superposition times for each training step.
When several training steps are combined, their backward gradients are recorded successively before the network parameters are updated together.
During superposition training, each feature channel is randomly assigned to a sub-sample, and the normal and superposition strategies are used concurrently for 700 steps.
For fine-tuning, the computational parameters $\Theta_{\mathcal C}$ are frozen and only the superposition parameters $\Theta_{\mathcal S}$ are updated. The fine-tuning steps use $t_{\text{super}}=1,2,3$, a learning rate of $0.00001$, and 100 iterations.
\begin{table}[htbp]
\centering
\caption{Forward and backward operations and parameters of the SDNO training strategies.}
\label{tab:training_strategy}
\resizebox{\textwidth}{!}{%
\begin{tabular}{lllll}
\toprule
Training step & Feed forward & Backward parameter & Learning rate & Superposition times \\ \midrule
Normal & $\hat{\mathcal C}(X)$ & $\Theta_{\mathcal C}$ & 0.0001 & --- \\
Superposition & $\hat{\mathcal S}(\hat{\mathcal C}(X^{(A)}),\hat{\mathcal C}(X^{(B)}))$ & $\Theta_{\mathcal C},\Theta_{\mathcal S}$ & 0.00001 & 1 \\
Fine-tuning & $\hat{\mathcal S}(\hat{\mathcal C}(X^{(A)}),\hat{\mathcal C}(X^{(B)}))$ & $\Theta_{\mathcal S}$ & 0.00001 & 1, 2, 3 \\
\bottomrule
\end{tabular}}
\end{table}
\section{Results and discussion}
\par
This section employs the SDNO proposed to predict the cooling temperature field of the gas film at the end wall of Pak-B.
This study incorporates two training schemes, superposition training and fine-tuning training, designed to enhance the prediction accuracy and generalization ability of the model.
The trained model will be compared with the fully supervised model employed in Section 3.2 to assess the effectiveness of the SDNO network in enhancing prediction performance.
\subsection{Joint Training Results of the Computational Network and the Superposition Network}
In alignment with the training strategy outlined in Section 4.2, this study employs the identical dataset as in Section 3.2 for model training to ensure evaluation consistency, ultimately completing the parameter training of both component-networks.
Initially, the computational network is tested independently. As previously mentioned, during superposition training, the information updated through reverse differentiation not only optimizes the superposition network $\hat{\mathcal{S}}(\dot)$ but also synchronously adjusts the parameters in the computational network $\hat{\mathcal{C}}(\dot)$.
This decomposition training strategy significantly enhances the diversity of sample permutations and combinations in the computational model, thereby enhancing the prediction accuracy itself.
\par
\begin{figure}[ht]
\begin{center}
\includegraphics[width=0.7\textwidth, trim = 0 0 0 0]{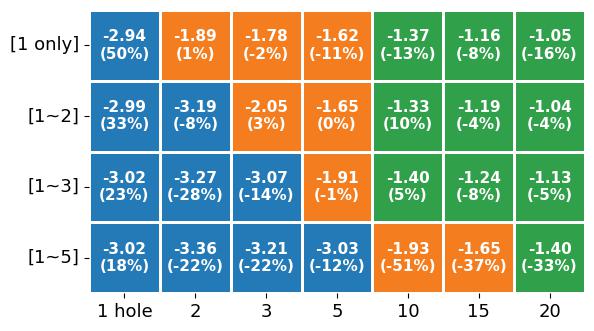}
\end{center}
\caption{The prediction error for ID and OOD samples of the computational network}
\label{heat_super_0}
\end{figure}

Figure \ref{heat_super_0} illustrates the prediction accuracy of various numbers of gas film pore samples in the validation set using the computational network $\hat{\mathcal{C}}(\dot)$ only.
Each row in the figure represents an individual model instance, with the number in square brackets on the left denoting the size of the sample set used by that model instance during training.
The numbers displayed in the squares are $\log_{10}e_{\mathrm{FAE}}$. Percentages in parentheses show the signed relative change $(e_{\mathrm{FAE}}-e_{\mathrm{FAE}}^{\mathrm{base}})/e_{\mathrm{FAE}}^{\mathrm{base}}$; negative percentages indicate lower error than the baseline.
\par
Through data analysis, it was observed that the training method of random decomposition significantly enriched the training diversity of samples.
This strategy not only improved the prediction accuracy of most samples (excluding samples with a single film-cooling hole) but also substantially enhanced the prediction of adaptability about OOD samples.
However, as evident from the color changes in Fig. \ref{heat_super_0}, although the model exhibits considerable accuracy improvement on samples within the training set, its prediction accuracy for samples beyond the training set distribution remains insufficient to meet engineering application requirements.
\par
Additionally, the superposition network $\hat{\mathcal{S}}(\dot)$ is also incorporated into the forward reasoning process to predict the sample.
When $\hat{\mathcal{S}}(\dot)$ is utilized, a sample undergoes the "decomposing-prediction-superposition" process, as depicted in Fig \ref{super_step}.
The choice of decomposition methods also influences the final predicting accuracy.
\par
\begin{figure}[ht]
\begin{center}
\includegraphics[width=0.7\textwidth, trim = 0 0 0 0]{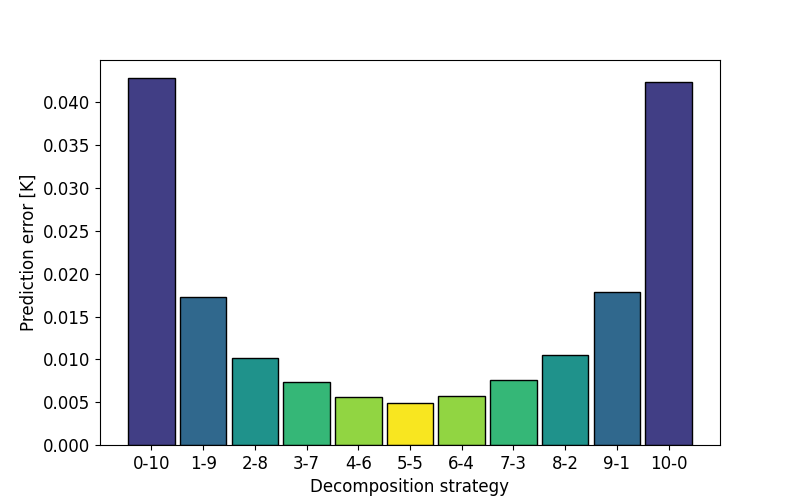}
\end{center}
\caption{The prediction error of 10-holes samples with different decomposition strategies}
\label{chart_bar}
\end{figure}

Unlike the completely random allocation of film-cooling holes to sub-samples in the training process to enhance training diversity, the reasoning process employs a fixed allocation strategy to enhance prediction accuracy.
Fig. \ref{chart_bar} displays the average temperature error of 100 samples in the validation set when a validation set of 10 gas film holes is decomposed into varying numbers of sub-samples.
The error magnitude follows a distinct pattern based on the decomposition strategy:
distributing gas film holes to different sub-samples in a relatively uniform manner minimizes the prediction errors for OOD samples, thereby effectively improving the final prediction accuracy.
Based on the above analysis, this work adopts a "homogenization" strategy to allocate gas film holes to each sub-sample when predicting.
\par
\begin{figure}[ht]
\begin{center}
\includegraphics[width=0.7\textwidth, trim = 0 0 0 0]{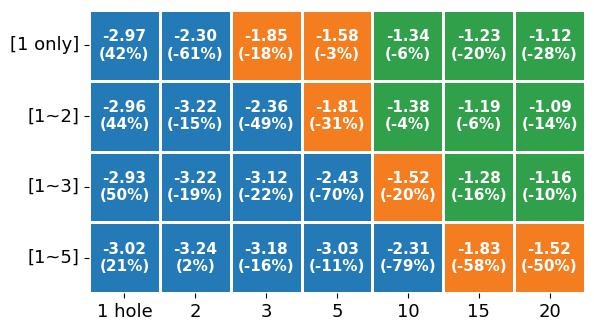}
\end{center}
\caption{The prediction error for ID and OOD samples of the SDNO network}
\label{heat_super_1}
\end{figure}
\begin{figure}[ht]
\begin{center}
\includegraphics[width=0.5\textwidth, trim = 0 0 0 0]{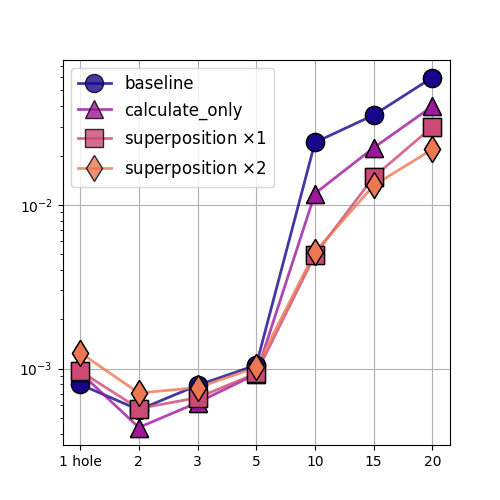}
\end{center}
\caption{The prediction error for ID and OOD samples of the SDNO network}
\label{line_super}
\end{figure}

Figure \ref{heat_super_1} displays the prediction accuracy when the superposition network $\hat{\mathcal{S}}(\dot)$ is invoked once, with the improvement relative to the baseline data shown in parentheses.
Following superposition, the model's extrapolation ability has been significantly enhanced, with samples featuring a number of air film holes within the one time superposition range (i.e., not exceeding twice the number of air film holes in the training set) achieving engineering substitution accuracy, with errors less than 0.0050K.
For instance, when the number of gas film holes contained in the training set samples is less than or equal to 5, the prediction accuracy for 10 film-cooling holes reached 0.0048K, essentially meeting engineering requirements.
Compared to fully supervised training, the SDNO network reduced the prediction error of the 10-hole samples by 79\% with the same training dataset.
Figure \ref{line_super} illustrates the specific prediction accuracy for the reference model and superposition model obtained by training the data of [1,2,3,5] film-cooling holes using the training set.
The superposition training strategy effectively enhances the accuracy of the computational network in predicting samples containing a complex number of film-cooling holes, albeit with a slight decrease in accuracy for single-sample models due to their inability to be decomposed.
When the superposition network is employed for inference prediction, the error of samples within the data set distribution is enlarged, while the error of samples outside the data set distribution is significantly reduced due to the superposition process.
This tendency is exacerbated when calculating 4 sub-samples and superimposing 2 times.
\par
\begin{figure}[htbp]
\centering
\subfigure[case A with 10 film cooling holes]{
\begin{minipage}[t]{0.9\linewidth}
\centering
\includegraphics[width=1\textwidth]{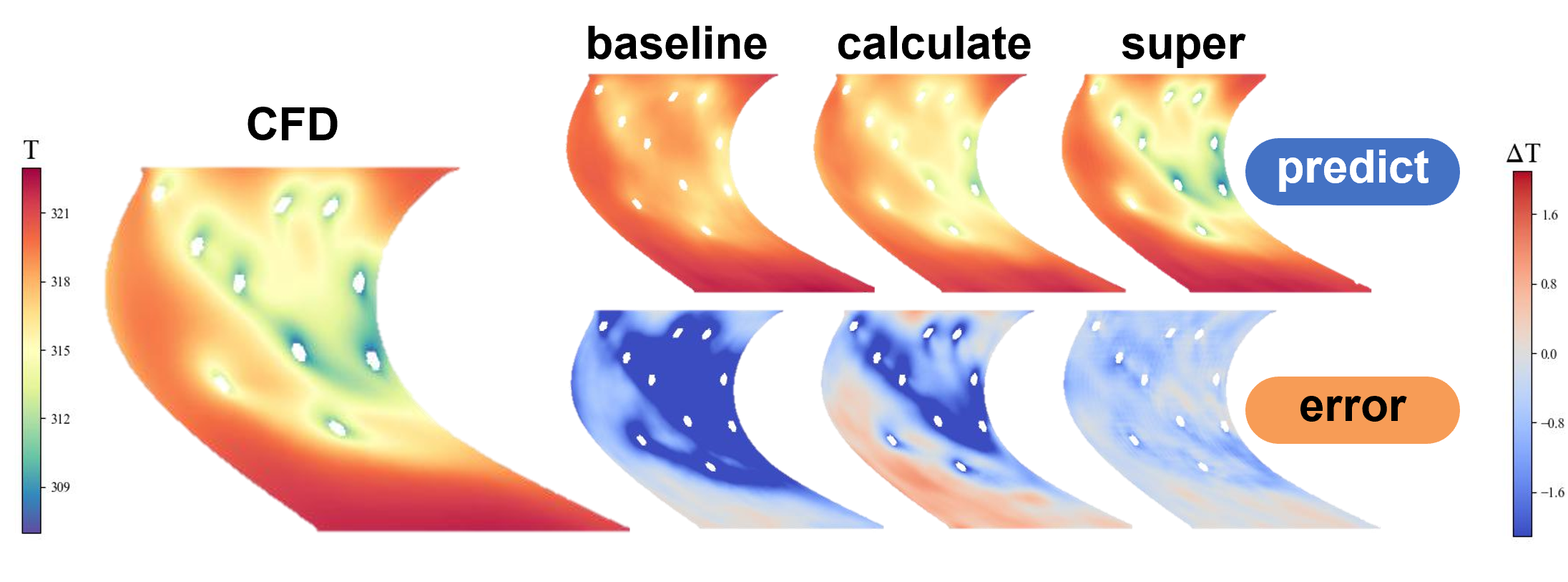}
\end{minipage}%
}%

\subfigure[case B with 10 film cooling holes]{
\begin{minipage}[t]{0.9\linewidth}
\centering
\includegraphics[width=1\textwidth]{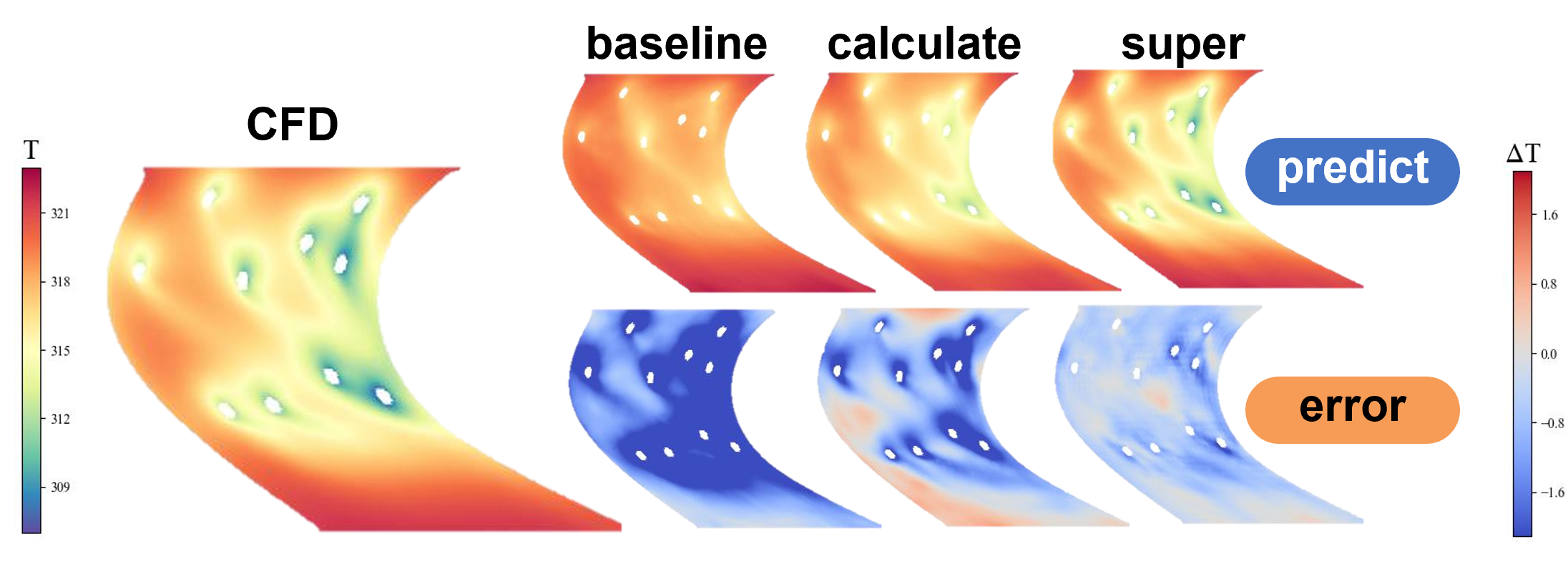}
\end{minipage}
}%

\subfigure[case C with 10 film cooling holes]{
\begin{minipage}[t]{0.9\linewidth}
\centering
\includegraphics[width=1\textwidth]{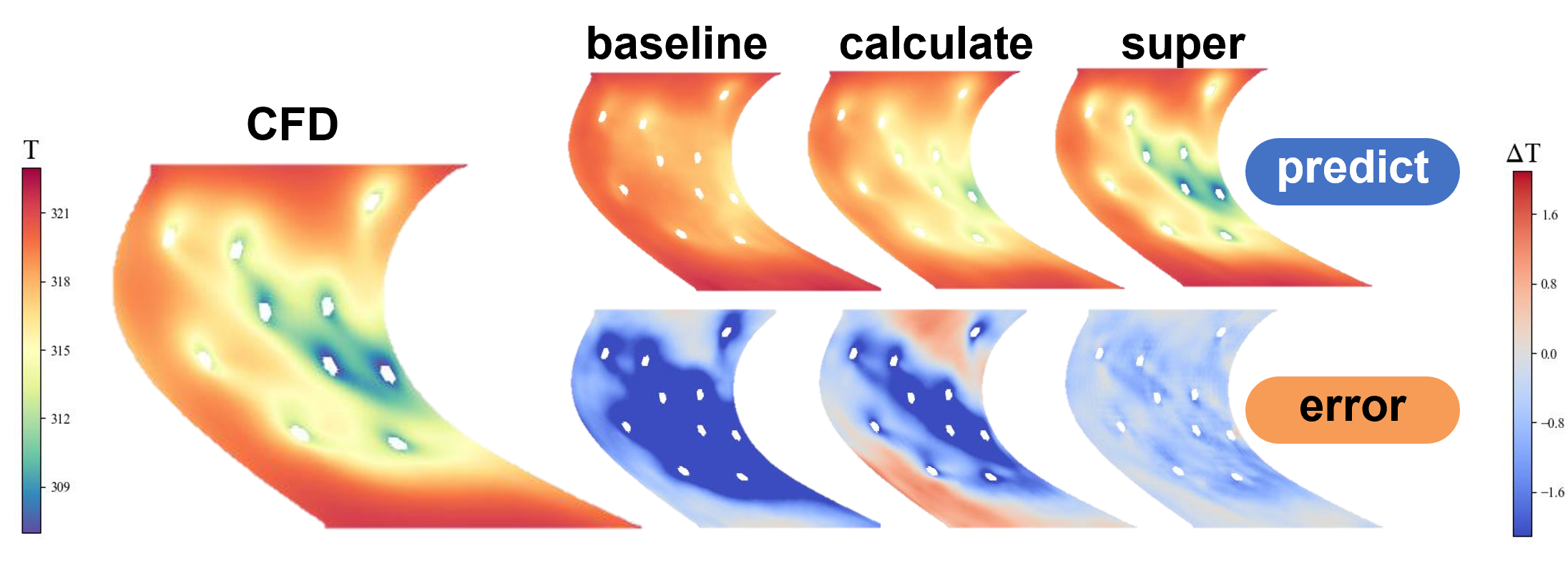}
\end{minipage}
}%
\centering
\caption{The comparison of temperature fields from CFD simulation and SDNO prediction.}
\label{cases_super_field}
\end{figure}

To demonstrate the extrapolation prediction ability of different models more specifically, 3 samples with 10 film-cooling holes are selected randomly from the valid dataset to present the predicted results.
Figure \ref{cases_super_field} intuitively demonstrates that the superimposed network proposed in this paper enhances the model's extrapolation ability.
The baseline model significantly overestimate the temperature of the end wall surface.
However, after utilizing the superposition network for inference prediction, the model accurately predicts the low temperature region formed by several film-cooling holes.
The maximum temperature error of the entire endwall surface is less than 0.5K.
\par
Upon examining the temperature field distribution details in Fig. \ref{cases_super_field}, it is evident that the SDNO network accurately predicts both the local low-temperature area in the upstream position of the film-cooling holes due to heat transfer in the film-cooling holes channel and the cold film area downstream of the film-cooling holes.
Furthermore, in Fig. \ref{cases_super_field}, near the suction surface, the influence ranges of several film-cooling holes are superimposed, forming a distinct low-temperature region where the temperature is significantly lower than the limitation of the temperature range caused by a single film-cooling hole.
Despite this porous coupled flow phenomenon, the operator network of the fusion superposition principle proposed in this paper achieves accurate capture.

\subsection{Fine-Tuning Training Results of the Superposition Network}
As summarized in Tables~\ref{tab:transformer} and~\ref{FNO}, the parameters of a superposition model $\hat{\mathcal{S}}(\dot)$ are significantly smaller in scale compared to those of a computational model $\hat{\mathcal{C}}(\dot)$, resulting in a lower computational cost for fixing the parameters $\Theta_{\mathcal{C}}$ and training the parameters $\Theta_{\mathcal{S}}$ alone.
\par
This article employs the concept of the pre-training and the fine-tuning task with low data scale and computational cost, aimed at further enhancing the prediction accuracy of the network.
The pre-trained model uses the 1-, 2-, 3-, and 5-hole training samples. Fine-tuning then uses the 10 available samples for each selected 10-, 15-, or 20-hole configuration; the training settings are given in Table~\ref{tab:training_strategy}.
During fine-tuning training, the superposition network is invoked once, twice, and three times, with the gradient parameters added and adjusted accordingly.%
\begin{figure}[htbp]
\centering
\subfigure[$t_{\text{super}}=1$ in inference]{
\begin{minipage}[t]{0.48\linewidth}
\centering
\includegraphics[width=1\textwidth]{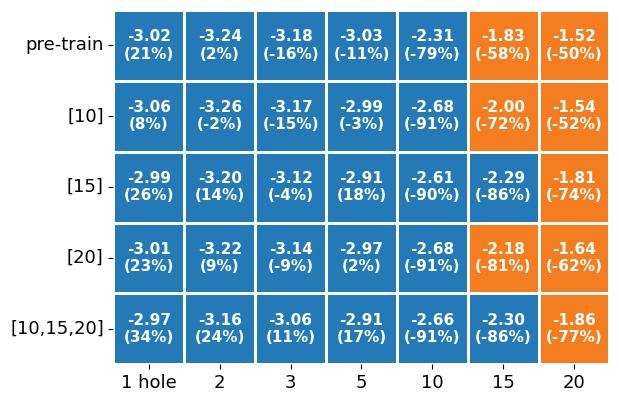}
\end{minipage}%
}%
\subfigure[$t_{\text{super}}=2$ in inference]{
\begin{minipage}[t]{0.48\linewidth}
\centering
\includegraphics[width=1\textwidth]{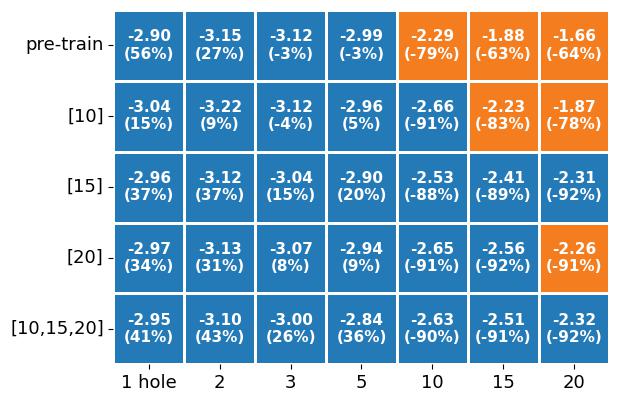}
\end{minipage}
}%
\centering
\caption{The prediction error of fine-tuning SDNO network with different training data}
\label{heatmap_ft}
\end{figure}
Figure \ref{heatmap_ft} depicts the error of each model for each result after fine-tuning,
where Fig. \ref{heatmap_ft}(a) and (b) display the results when the superposition time $t_{\text{super}}$ set as 1 and 2, respectively.
It is evident that the model's extrapolation ability in prediction significantly improves as $t_{\text{super}}$ increases.
\par
\begin{figure}[ht]
\begin{center}
\includegraphics[width=0.5\textwidth, trim = 0 0 0 0]{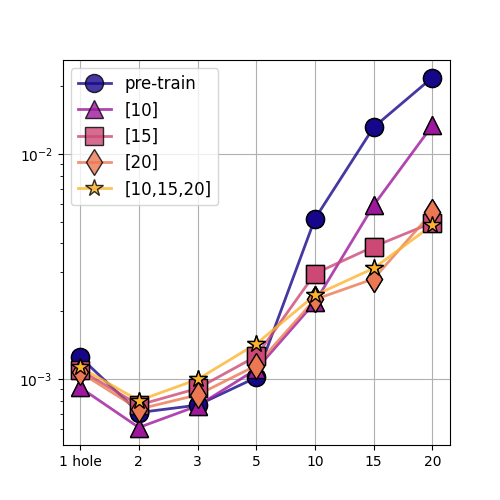}
\end{center}
\caption{The prediction error of fine-tuning SDNO network with different training data}
\label{line_ft}
\end{figure}

The precision changes of the model fine-tuned with different samples, with the $t_{\text{super}}$ as 2, were illustrated as a broken line in Fig. \ref{line_ft}.
It is evident that, compared with the pre-trained model, the prediction accuracy of the samples within the training set slightly decreased after fine-tuning, whereas the prediction accuracy of the samples outside the distribution notably improved.
In comparison with the samples used for fine-tuning, it is observed that the prediction accuracy of the 15-holes and 20-holes samples slightly improves when only 10-holes samples are used for fine-tuning. Conversely, the prediction accuracy of all samples outside the distribution significantly improves when only 20-holes samples are used for fine-tuning.
\par
Additionally, it is noteworthy that the results of fine-tuning with 20-holes samples outperform those with 10-holes samples.
The presence of a superposition network enables the decomposition of samples downwards, with 20-holes samples capable of decomposing $C_{20}^{10}$ different 10-holes samples.
Fine-tuning serves as a generalization for improving prediction accuracy.
For instance, when only 10 20-holes samples are added for fine-tuning, the prediction accuracy of 10-holes samples also sees a significant improvement.

\par
The mean relative variance error provides a complementary measure of spatial error variation:
\begin{equation}
 e_{\mathrm{MRVE}}=\frac{1}{N_{\mathrm{valid}}}
 \sum_{i=1}^{N_{\mathrm{valid}}}
 \frac{\operatorname{Var}(f_{b,i}-\widehat f_{b,i})}
      {\operatorname{Var}(f_{b,i})}.
 \label{eq:mrve}
\end{equation}
Here $f_{b,i}$ and $\widehat f_{b,i}$ are the reference and predicted temperature fields for validation sample $i$, and the variance is evaluated over spatial points within that sample. This dimensionless error measures the variance of the prediction error relative to the spatial variance of the reference field, whereas $e_{\mathrm{FAE}}$ in Eq.~\ref{eq:fae} measures absolute temperature error.
Table~\ref{tab:mrve} reports $e_{\mathrm{MRVE}}$ for the baseline and SDNO variants pre-trained with samples containing 1, 2, 3, and 5 holes.
The baseline MRVE rises to 41.884\%, 37.297\%, and 33.697\% for 10-, 15-, and 20-hole validation samples. With fine-tuning and three superposition levels, these errors are 1.084\%, 1.383\%, and 2.756\%, respectively. The relative MRVE reduction therefore exceeds 90\% for each of these three OOD configurations. This statement concerns the multi-level fine-tuned model; it does not imply that every model row in Table~\ref{tab:mrve} achieves a 90\% reduction.
\begin{table}[htbp]
\centering
\caption{Mean relative variance error of the SDNO and baseline models for validation datasets with different numbers of cooling holes. All entries are percentages.}
\label{tab:mrve}
\resizebox{\textwidth}{!}{%
\begin{tabular}{lrrrrrrr}
\toprule
Network & 1 hole & 2 holes & 3 holes & 5 holes & 10 holes & 15 holes & 20 holes \\ \midrule
Baseline & 0.868\% & 0.352\% & 0.442\% & 0.432\% & 41.884\% & 37.297\% & 33.697\% \\
Pre-train: Calculate only & 0.773\% & 0.181\% & 0.237\% & 0.464\% & 18.714\% & 16.221\% & 22.686\% \\
Pre-train: Super $\times$ 1 & 1.001\% & 0.364\% & 0.281\% & 0.447\% & 1.405\% & 5.506\% & 12.564\% \\
Pre-train: Super $\times$ 2 & 1.416\% & 0.464\% & 0.335\% & 0.481\% & 1.478\% & 3.727\% & 7.311\% \\
Fine-tuning: Super $\times$ 1 & 1.251\% & 0.445\% & 0.409\% & 0.639\% & 0.942\% & 2.364\% & 11.379\% \\
Fine-tuning: Super $\times$ 2 & 1.535\% & 0.571\% & 0.511\% & 0.743\% & 1.062\% & 1.391\% & 2.794\% \\
Fine-tuning: Super $\times$ 3 & 1.828\% & 0.714\% & 0.577\% & 0.782\% & 1.084\% & 1.383\% & 2.756\% \\
\bottomrule
\end{tabular}}
\end{table}

\begin{figure}[htbp]
\centering
\subfigure[case A with 10 film cooling holes]{
\begin{minipage}[t]{0.48\linewidth}
\centering
\includegraphics[width=1\textwidth]{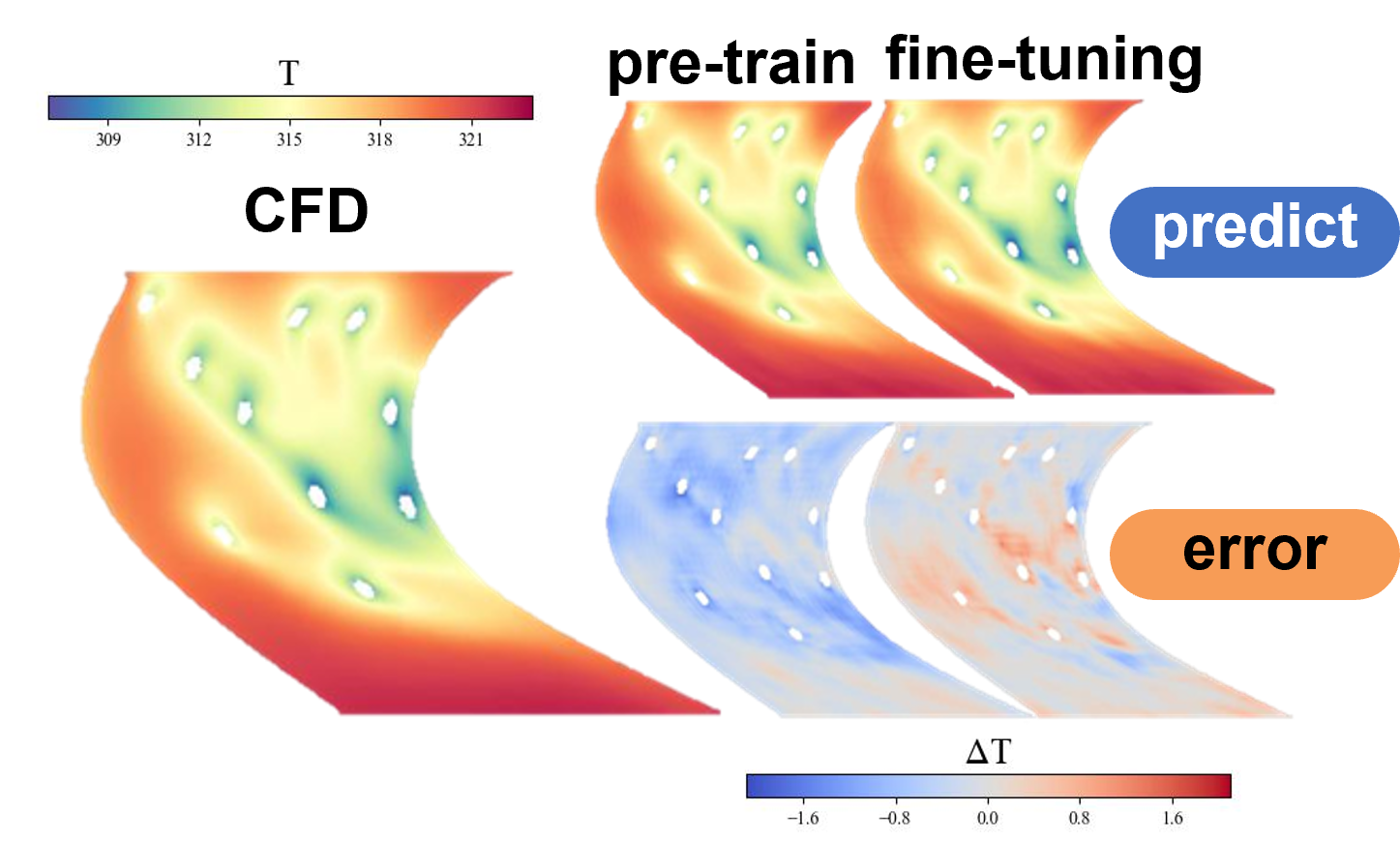}
\end{minipage}%
}%
\subfigure[case B with 10 film cooling holes]{
\begin{minipage}[t]{0.48\linewidth}
\centering
\includegraphics[width=1\textwidth]{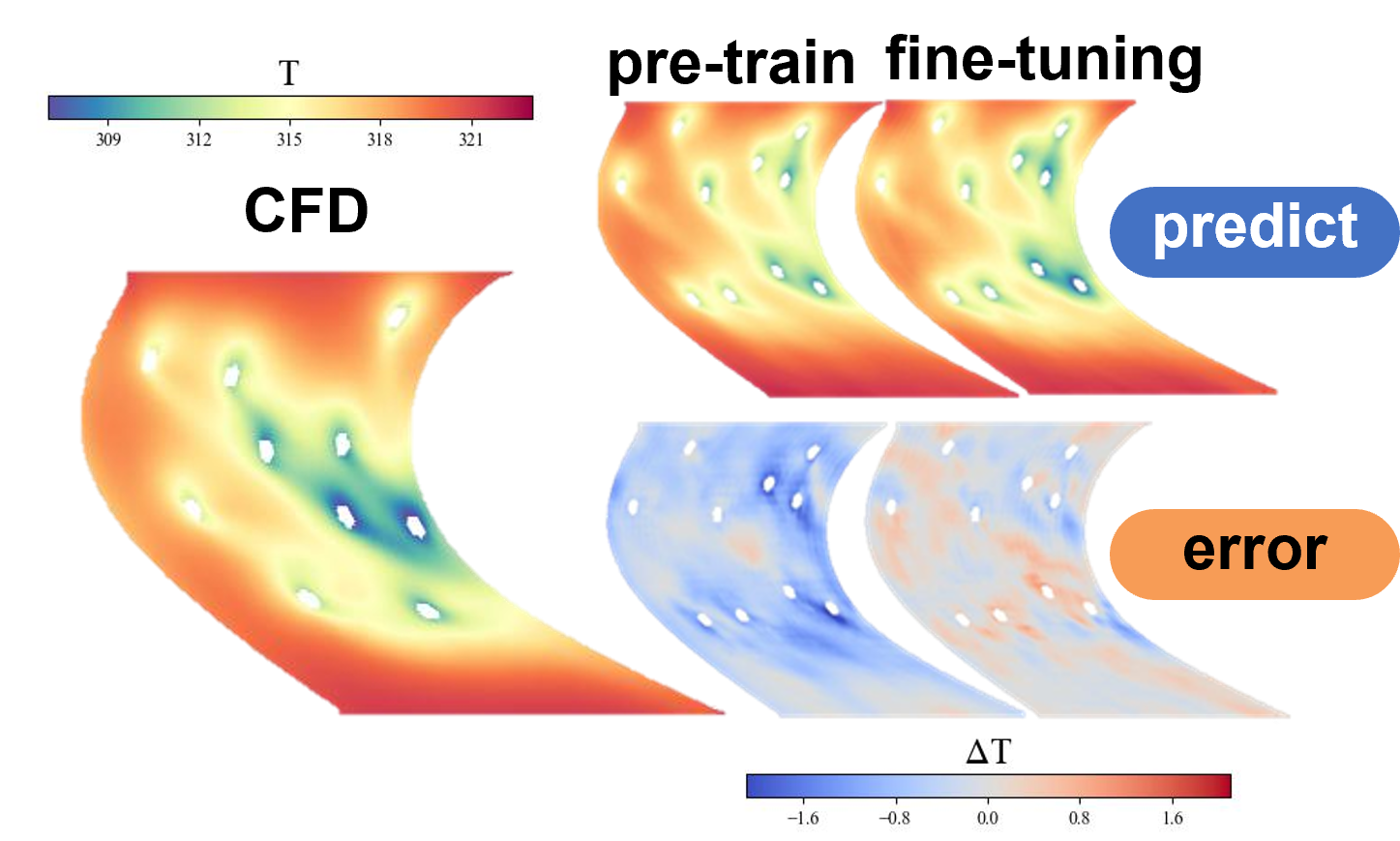}
\end{minipage}
}%

\subfigure[case A with 15 film cooling holes]{
\begin{minipage}[t]{0.48\linewidth}
\centering
\includegraphics[width=1\textwidth]{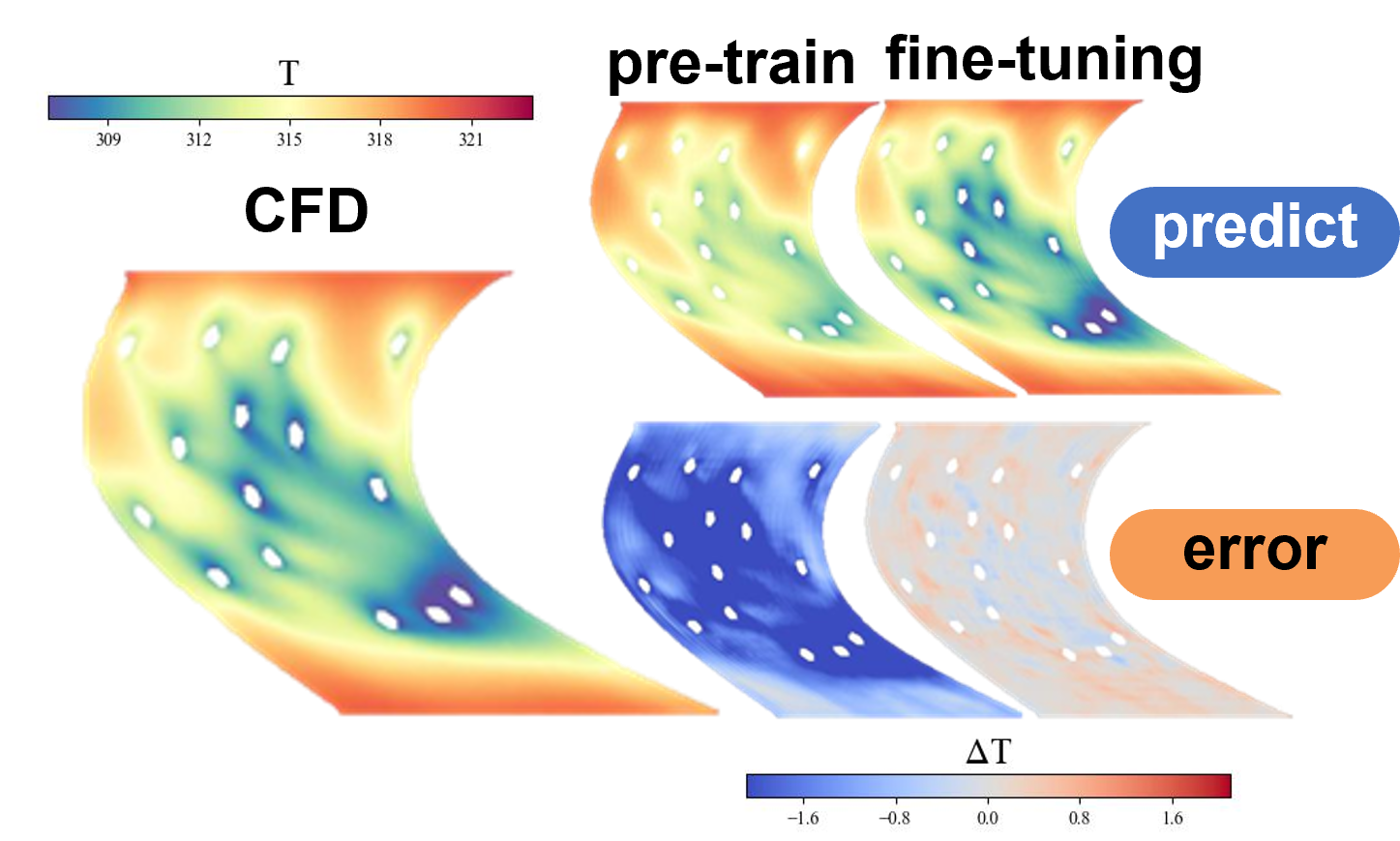}
\end{minipage}%
}%
\subfigure[case B with 15 film cooling holes]{
\begin{minipage}[t]{0.48\linewidth}
\centering
\includegraphics[width=1\textwidth]{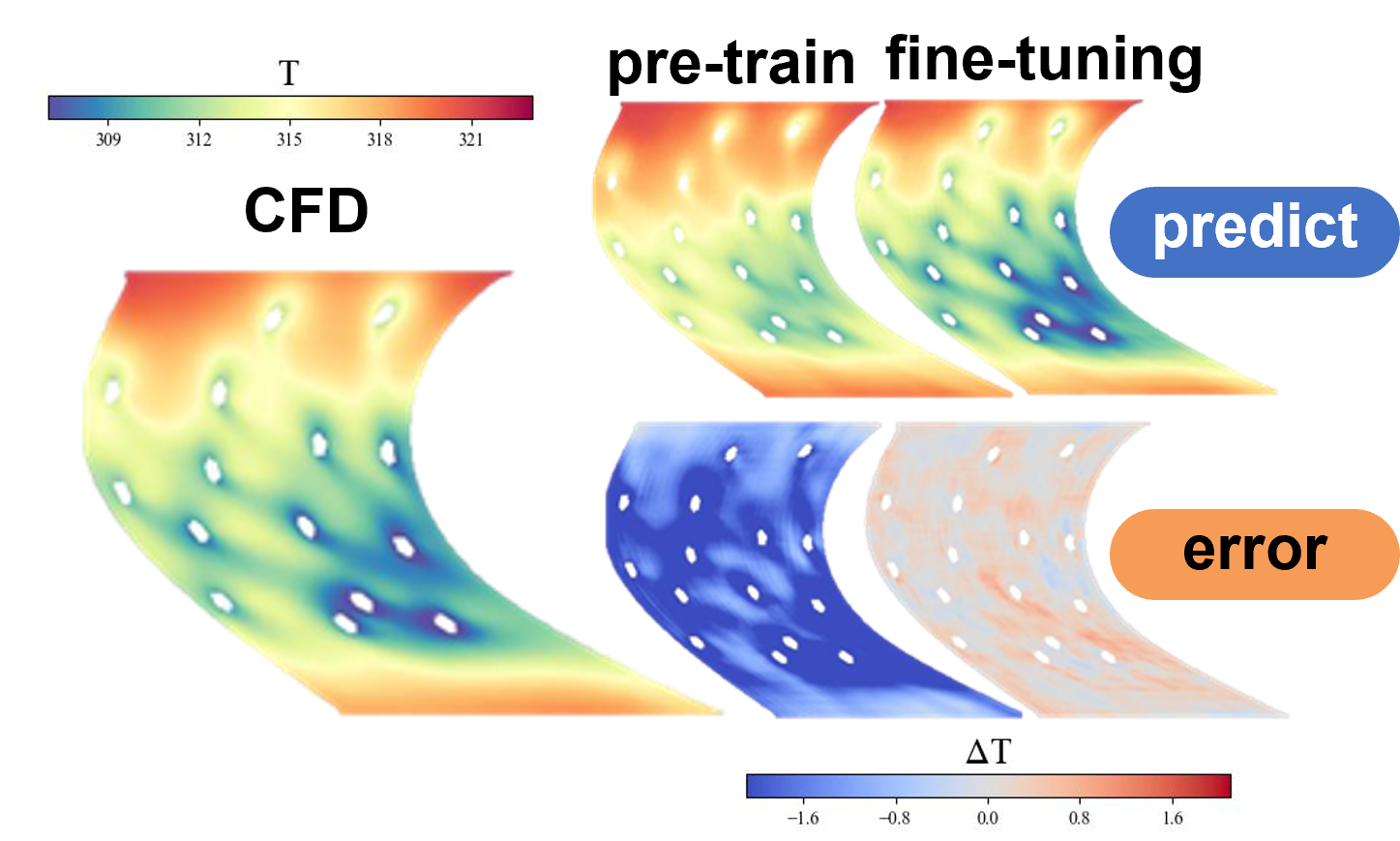}
\end{minipage}
}%

\subfigure[case A with 20 film cooling holes]{
\begin{minipage}[t]{0.48\linewidth}
\centering
\includegraphics[width=1\textwidth]{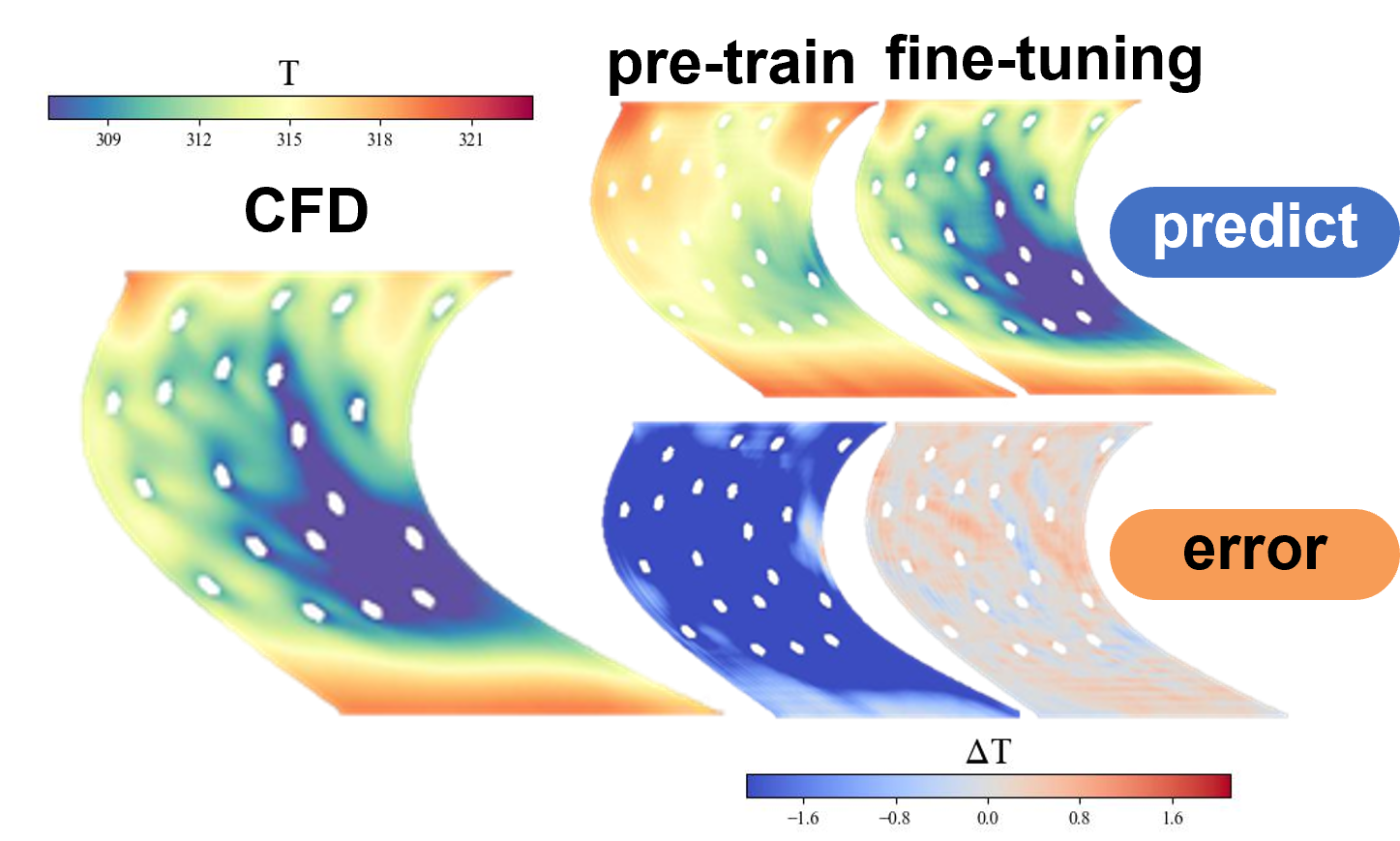}
\end{minipage}%
}%
\subfigure[case B with 20 film cooling holes]{
\begin{minipage}[t]{0.48\linewidth}
\centering
\includegraphics[width=1\textwidth]{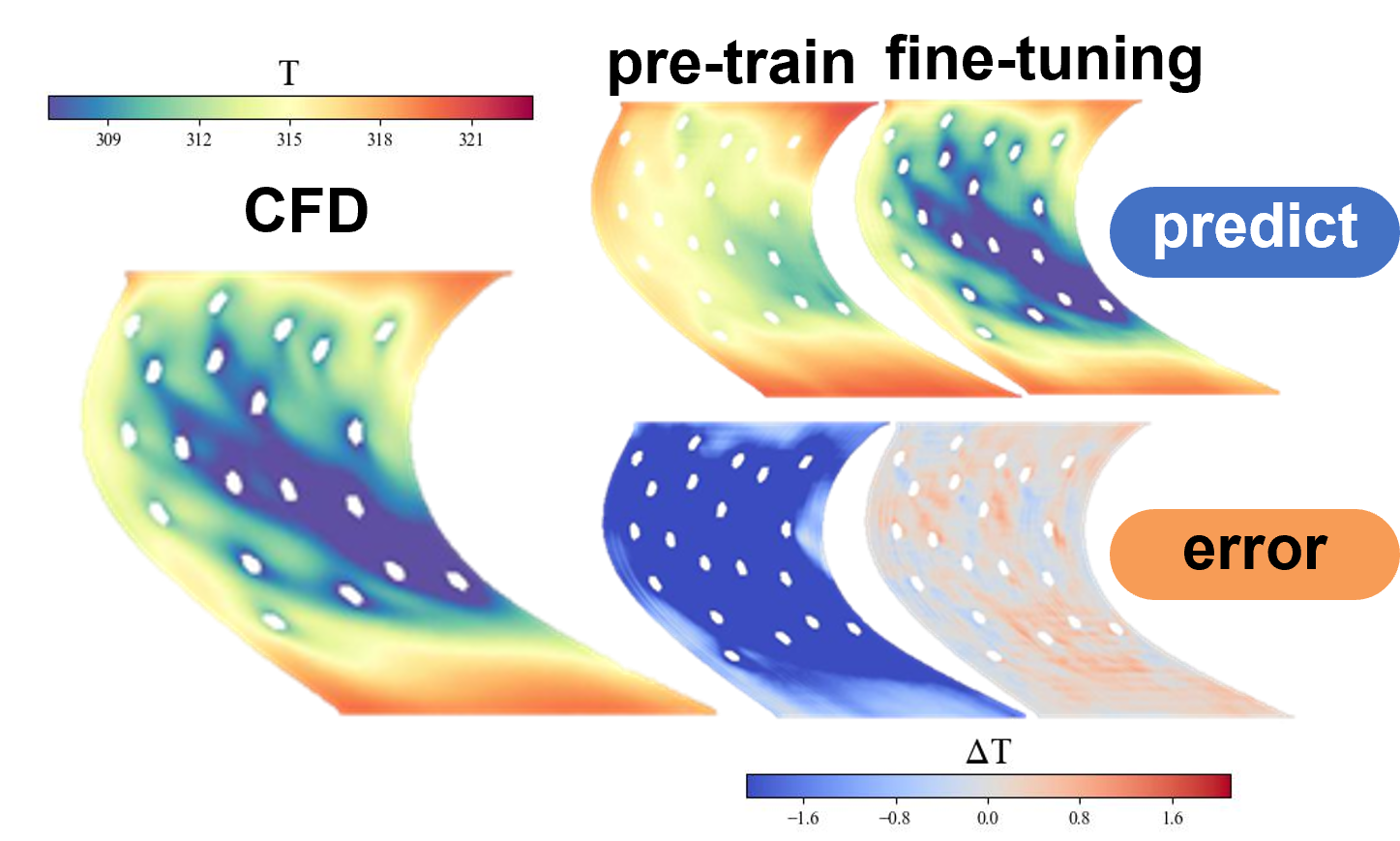}
\end{minipage}
}%

\centering
\caption{The comparison of temperature fields from CFD simulation and fine-tuned SDNO prediction.}
\label{case_fields_ft}
\end{figure}

Following the approach outlined in Section 5.1, random samples are selected from the validation dataset to provide a more intuitive demonstration of the effects of fine-tuning training.
Figure \ref{case_fields_ft} illustrates the prediction results for two samples with 10, 15, and 20 holes, respectively.
It is evident that the extrapolation capability of the fine-tuned model has significantly improved. The pre-trained model often overestimates the temperature field of samples beyond the training distribution due to the absence of low-temperature regions influenced by combinations of pores in the training set as the number of holes increases.
This difficulty in correctly evaluating the degree of temperature drop under superposition is addressed through a small amount of fine-tuning sample training, resulting in a maximum temperature field prediction error of less than 0.5K for each sample.
\par
\begin{figure}[htbp]
\centering
\subfigure[case A with 10 film cooling holes]{
\begin{minipage}[t]{0.8\linewidth}
\centering
\includegraphics[width=1\textwidth]{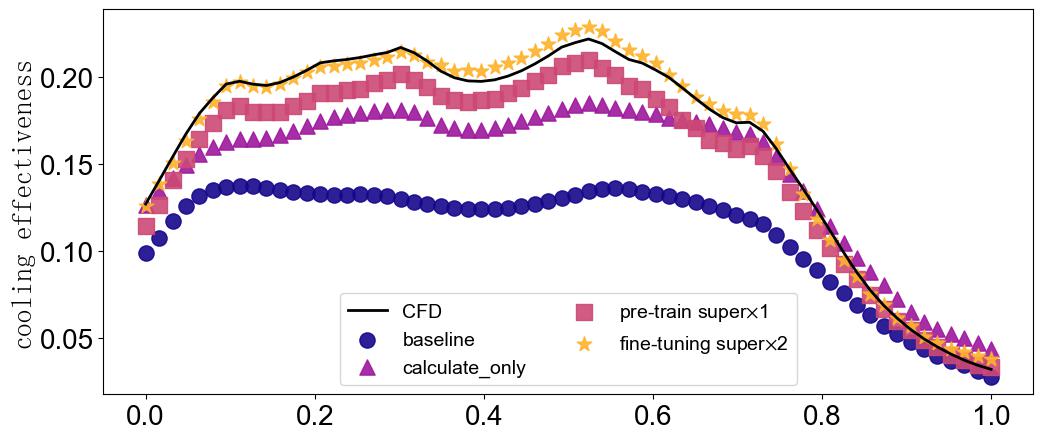}
\end{minipage}%
}%

\subfigure[case B with 10 film cooling holes]{
\begin{minipage}[t]{0.8\linewidth}
\centering
\includegraphics[width=1\textwidth]{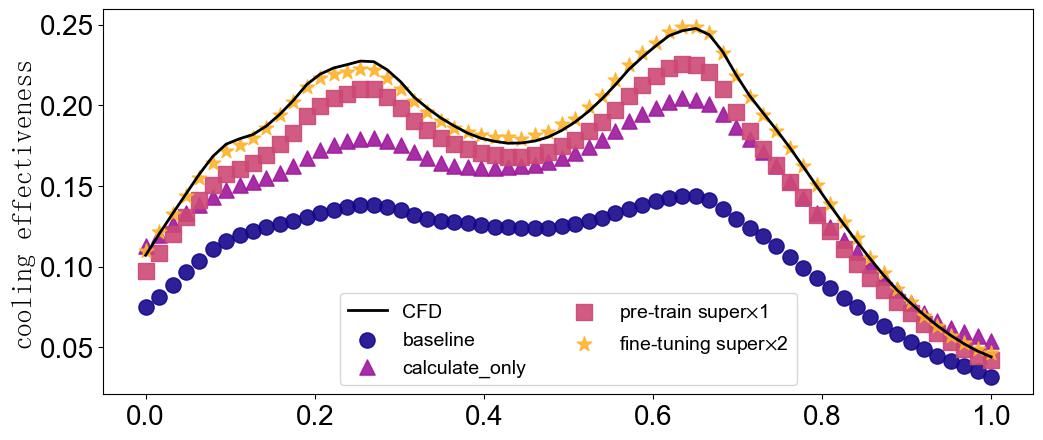}
\end{minipage}
}%

\subfigure[case C with 10 film cooling holes]{
\begin{minipage}[t]{0.8\linewidth}
\centering
\includegraphics[width=1\textwidth]{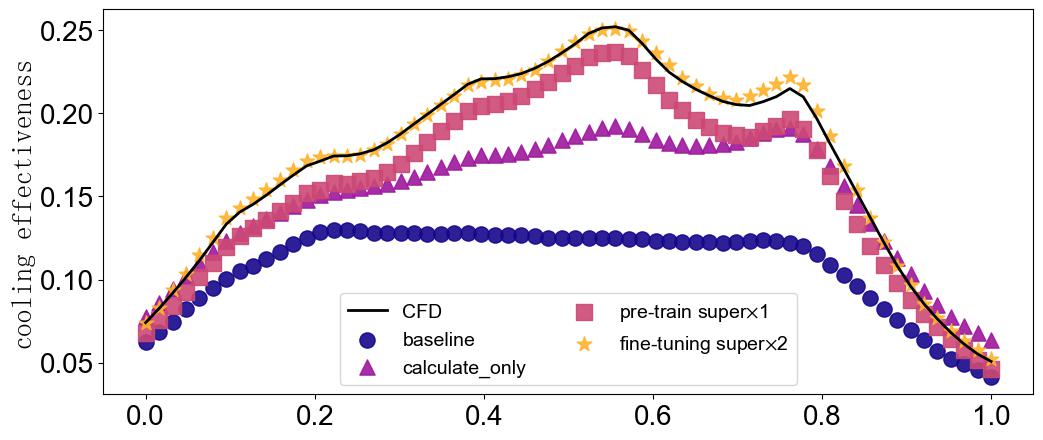}
\end{minipage}
}%
\centering
\caption{The comparison of axial-average cooling effectiveness from CFD simulation and the prediction models.}
\label{case_curves}
\end{figure}

Figure \ref{case_curves} further illustrates the temperature cloud map and the axial average gas film effectiveness curve before and after superposition, using the three 10-hole cases A, B, and C shown above. The figure depicts the accuracy of the basic model, the calculation model, the superposition model, and the fine-tuning model in extrapolation prediction. It is evident that the basic model fails to accurately predict out-of-distribution samples, while the accuracy of the calculation model significantly improves after training. However, accurately estimating regions with high gas film effectiveness in cold air concentration remains challenging (as depicted in Fig.~\ref{case_curves}(c) at the relative axial position 0.6). In contrast, the superimposed model can generally estimate the trend of effectiveness distribution along the axis with only slight deviations in magnitude. The final fine-tuning process eliminates this bias, enabling accurate prediction of the effectiveness curve.

\begin{figure}[htbp]
\centering
\subfigure[Average film cooling effectiveness]{
\begin{minipage}[t]{0.48\linewidth}
\centering
\includegraphics[width=1\textwidth]{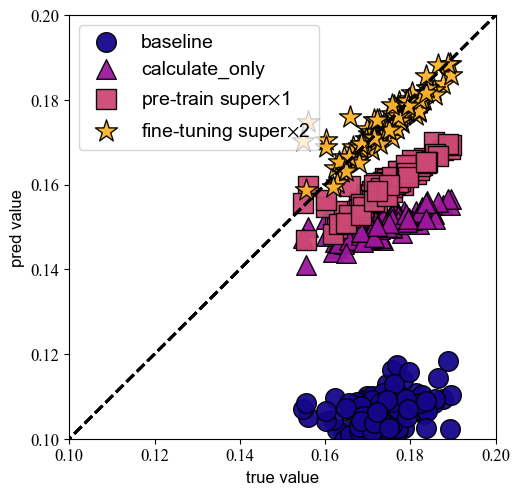}
\end{minipage}%
}%
\subfigure[Overheating areas]{
\begin{minipage}[t]{0.48\linewidth}
\centering
\includegraphics[width=0.99\textwidth]{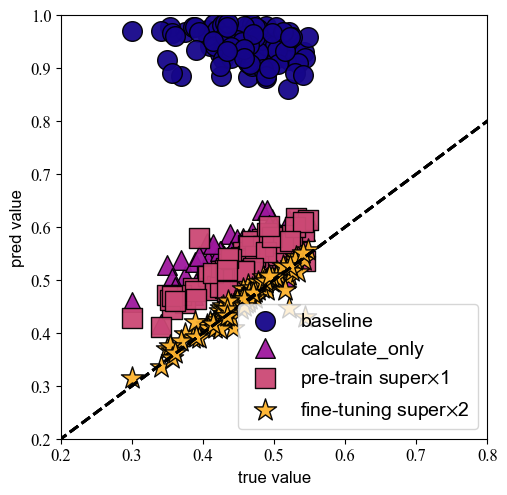}
\end{minipage}
}%
\centering
\caption{The comparison between the simulation and prediction performace with different deep learning networks}
\label{performance}
\end{figure}

Finally, the validation set of 10-hole layouts is used to compare scalar performance measures in Fig.~\ref{performance}. The area-averaged film cooling effectiveness is
\[
 \phi=\frac{T_\infty-\int_\Omega T(x)\,\mathrm{d}s/\int_\Omega\mathrm{d}s}{T_\infty-T_c}.
\]
The overheating area is $A_h=\int_{\{x\in\Omega:\eta(x)<0.15\}}\mathrm{d}s$, with the local effectiveness $\eta(x)=(T_\infty-T(x))/(T_\infty-T_c)$. Figure~\ref{performance}(b) compares the reference and predicted overheating areas.
The Fig. \ref{performance} vividly illustrates the accuracy enhancement from the baseline model to the calculation model, then to the superposition inference, and finally to the fine-tuning model.

\section{Conclusions}
This study integrates the film-cooling superposition principle with a deep neural operator to predict the temperature field on a Pak-B blade endwall. SDF preprocessing represents small cooling-hole geometries, a Transformer predicts sub-layout temperature fields, and a local FNO combines them through trainable superposition. The model is pre-trained with layouts containing 1, 2, 3, and 5 holes and evaluated on layouts containing up to 20 holes.
The superposition model reaches an average absolute temperature error of 0.0048 K for 10-hole validation samples, a 79\% reduction relative to the fully supervised baseline. Fine-tuning with a small additional dataset further improves extrapolation. With three superposition levels, the MRVE values for 10-, 15-, and 20-hole samples are 1.084\%, 1.383\%, and 2.756\%, compared with baseline values of 41.884\%, 37.297\%, and 33.697\%. Each of these OOD configurations therefore achieves more than 90\% relative MRVE reduction.
Although the approach reduces sample requirements, training still relies on high-fidelity CFD data. Future work can combine multi-fidelity data and additional physical constraints, and extend the input space to operating conditions, hole diameter, and injection angle. These extensions would support assessment of more varied turbomachinery cooling layouts.

\section*{Funding}
This study was supported by the National Science and Technology Major Project (Grant No. 2019-II-0008-0028).
\section*{Conflict of interest}
The authors declare that they have no conflict of interest.
\section*{Data availability statement}
The data that support the findings of this study are available from the corresponding author upon reasonable request.

\bibliographystyle{elsarticle-num-names}
\bibliography{bibfile}%

\begin{thebibliography}{16}
\expandafter\ifx\csname natexlab\endcsname\relax\def\natexlab#1{#1}\fi
\providecommand{\url}[1]{\texttt{#1}}
\providecommand{\href}[2]{#2}
\providecommand{\path}[1]{#1}
\providecommand{\DOIprefix}{doi:}
\providecommand{\ArXivprefix}{arXiv:}
\providecommand{\URLprefix}{URL: }
\providecommand{\Pubmedprefix}{pmid:}
\providecommand{\doi}[1]{\href{http://dx.doi.org/#1}{\path{#1}}}
\providecommand{\Pubmed}[1]{\href{pmid:#1}{\path{#1}}}
\providecommand{\bibinfo}[2]{#2}
\ifx\xfnm\relax \def\xfnm[#1]{\unskip,\space#1}\fi
\bibitem[{Yang et~al.(2020)Yang, Wang, Huang, and
  Rao}]{yangEstablishmentLongshorttermmemoryModel2020a}
\bibinfo{author}{L.~Yang}, \bibinfo{author}{Q.~Wang},
  \bibinfo{author}{K.~Huang}, \bibinfo{author}{Y.~Rao},
\newblock \bibinfo{title}{Establishment of a long-short-term-memory model to
  predict film cooling effectiveness under superposition conditions},
\newblock \bibinfo{journal}{International Journal of Heat and Mass Transfer}
  \bibinfo{volume}{160} (\bibinfo{year}{2020}) \bibinfo{pages}{120231}.
  \DOIprefix\doi{10.1016/j.ijheatmasstransfer.2020.120231}.
\bibitem[{Zhi et~al.(2023)Zhi, Li, Wen, Su, and
  Wang}]{zhiResearchPerformancePredictions2023}
\bibinfo{author}{R.~Zhi}, \bibinfo{author}{Z.~Li}, \bibinfo{author}{F.~Wen},
  \bibinfo{author}{L.~Su}, \bibinfo{author}{S.~Wang},
\newblock \bibinfo{title}{Research on performance predictions using single-hole
  film cooling based on {{PointNet}}},
\newblock \bibinfo{journal}{Physics of Fluids} \bibinfo{volume}{35}
  (\bibinfo{year}{2023}) \bibinfo{pages}{025108}.
  \DOIprefix\doi{10.1063/5.0136442}.
\bibitem[{Xing et~al.(2022)Xing, Luo, Du, and
  Wang}]{xingDirectInverseModel2022}
\bibinfo{author}{H.~Xing}, \bibinfo{author}{L.~Luo}, \bibinfo{author}{W.~Du},
  \bibinfo{author}{S.~Wang},
\newblock \bibinfo{title}{Direct and {{Inverse Model}} for {{Single-Hole Film
  Cooling With Machine Learning}}},
\newblock \bibinfo{journal}{Journal of Turbomachinery} \bibinfo{volume}{144}
  (\bibinfo{year}{2022}) \bibinfo{pages}{041006}.
  \DOIprefix\doi{10.1115/1.4052601}.
\bibitem[{Wang et~al.(2023{\natexlab{a}})Wang, Qian, Sun, Wang, and
  Cui}]{wangFastPredictionUncertainty2023}
\bibinfo{author}{Y.~Wang}, \bibinfo{author}{S.~Qian}, \bibinfo{author}{Y.~Sun},
  \bibinfo{author}{W.~Wang}, \bibinfo{author}{J.~Cui},
\newblock \bibinfo{title}{Fast prediction and uncertainty analysis of film
  cooling with a semi-sphere vortex generator using artificial neural network},
\newblock \bibinfo{journal}{AIP Advances} \bibinfo{volume}{13}
  (\bibinfo{year}{2023}{\natexlab{a}}) \bibinfo{pages}{015303}.
  \DOIprefix\doi{10.1063/5.0132989}.
\bibitem[{Wang et~al.(2023{\natexlab{b}})Wang, Wang, Wang, Tao, Shen, and
  Cui}]{wangTwodimensionalPredictionSuperposition2023}
\bibinfo{author}{Y.~Wang}, \bibinfo{author}{Z.~Wang},
  \bibinfo{author}{W.~Wang}, \bibinfo{author}{G.~Tao},
  \bibinfo{author}{W.~Shen}, \bibinfo{author}{J.~Cui},
\newblock \bibinfo{title}{Two-dimensional prediction of the superposition film
  cooling with trench based on conditional generative adversarial network},
\newblock \bibinfo{journal}{International Journal of Thermal Sciences}
  \bibinfo{volume}{184} (\bibinfo{year}{2023}{\natexlab{b}})
  \bibinfo{pages}{107976}. \DOIprefix\doi{10.1016/j.ijthermalsci.2022.107976}.
\bibitem[{Li et~al.(2022)Li, Su, Wen, Zeng, Wang, and
  Zhang}]{liDeepLearningMethod2022}
\bibinfo{author}{Z.~Li}, \bibinfo{author}{L.~Su}, \bibinfo{author}{F.~Wen},
  \bibinfo{author}{J.~Zeng}, \bibinfo{author}{S.~Wang},
  \bibinfo{author}{J.~Zhang},
\newblock \bibinfo{title}{Deep learning method for fast prediction of film
  cooling performance},
\newblock \bibinfo{journal}{Physics of Fluids} \bibinfo{volume}{34}
  (\bibinfo{year}{2022}) \bibinfo{pages}{047111}.
  \DOIprefix\doi{10.1063/5.0087727}.
\bibitem[{Wang et~al.(2024)Wang, Lou, Li, and
  Yang}]{wangMetafourierNeuralOperators2024}
\bibinfo{author}{Q.~Wang}, \bibinfo{author}{J.~Lou}, \bibinfo{author}{Y.~Li},
  \bibinfo{author}{L.~Yang},
\newblock \bibinfo{title}{Meta-fourier neural operators for multi-task modeling
  of film cooling in gas turbine endwalls},
\newblock \bibinfo{journal}{Engineering Applications of Artificial
  Intelligence} \bibinfo{volume}{131} (\bibinfo{year}{2024})
  \bibinfo{pages}{107858}. \DOIprefix\doi{10.1016/j.engappai.2024.107858}.
\bibitem[{Zhang et~al.(2024)Zhang, Liu, Song, Lu, and
  Feng}]{zhangInvestigationImplementationGenerating2024}
\bibinfo{author}{W.~Zhang}, \bibinfo{author}{Z.~Liu},
  \bibinfo{author}{Y.~Song}, \bibinfo{author}{Y.~Lu},
  \bibinfo{author}{Z.~Feng},
\newblock \bibinfo{title}{An investigation on implementation of generating
  adversarial network-based surrogate models for prediction of turbine endwall
  film cooling effectiveness},
\newblock \bibinfo{journal}{Engineering Applications of Artificial
  Intelligence} \bibinfo{volume}{133} (\bibinfo{year}{2024})
  \bibinfo{pages}{108268}. \DOIprefix\doi{10.1016/j.engappai.2024.108268}.
\bibitem[{Wang et~al.(2024)Wang, Song, Guo, Li, and
  Feng}]{wangNOVELMULTIFIDELITYSURROGATE2023b}
\bibinfo{author}{Q.~Wang}, \bibinfo{author}{L.~Song}, \bibinfo{author}{Z.~Guo},
  \bibinfo{author}{J.~Li}, \bibinfo{author}{Z.~Feng},
\newblock \bibinfo{title}{A novel multi-fidelity surrogate for efficient
  turbine design optimization},
\newblock \bibinfo{journal}{Journal of Turbomachinery} \bibinfo{volume}{146}
  (\bibinfo{year}{2024}) \bibinfo{pages}{041011}.
  \DOIprefix\doi{10.1115/1.4064228}.
\bibitem[{Eberly(2006)}]{eberlyDistancePointEllipse}
\bibinfo{author}{D.~Eberly}, \bibinfo{title}{Distance from a {{Point}} to an
  {{Ellipse}}, an {{Ellipsoid}}, or a {{Hyperellipsoid}}},
  \bibinfo{howpublished}{Geometric Tools, LLC}, \bibinfo{year}{2006}.
\bibitem[{Sellers(1963)}]{sellersGASEOUSFILMCOOLING1963}
\bibinfo{author}{J.~P. Sellers},
\newblock \bibinfo{title}{{{GASEOUS FILM COOLING WITH MULTIPLE INJECTION
  STATIONS}}},
\newblock \bibinfo{journal}{AIAA Journal} \bibinfo{volume}{1}
  (\bibinfo{year}{1963}) \bibinfo{pages}{2154--2156}.
  \DOIprefix\doi{10.2514/3.2014}.
\bibitem[{Sasaki et~al.(1979)Sasaki, Takahara, Kumagai, and
  Hamano}]{sasakiFilmCoolingEffectiveness1979}
\bibinfo{author}{M.~Sasaki}, \bibinfo{author}{K.~Takahara},
  \bibinfo{author}{T.~Kumagai}, \bibinfo{author}{M.~Hamano},
\newblock \bibinfo{title}{Film {{Cooling Effectiveness}} for {{Injection}} from
  {{Multirow Holes}}},
\newblock \bibinfo{journal}{Journal of Engineering for Power}
  \bibinfo{volume}{101} (\bibinfo{year}{1979}) \bibinfo{pages}{101--108}.
  \DOIprefix\doi{10.1115/1.3446430}.
\bibitem[{Chen et~al.(2021)Chen, Mao, Hu, Su, and
  Yuan}]{chen2DPredictionMethod2021}
\bibinfo{author}{Z.~Chen}, \bibinfo{author}{Y.~Mao}, \bibinfo{author}{K.~Hu},
  \bibinfo{author}{X.~Su}, \bibinfo{author}{X.~Yuan},
\newblock \bibinfo{title}{2-{{D}} prediction method for multi-row film cooling
  effectiveness},
\newblock \bibinfo{journal}{Applied Thermal Engineering} \bibinfo{volume}{199}
  (\bibinfo{year}{2021}) \bibinfo{pages}{117607}.
  \DOIprefix\doi{10.1016/j.applthermaleng.2021.117607}.
\bibitem[{Andreini et~al.(2005)Andreini, Carcasci, Gori, and
  Surace}]{andreiniFilmCoolingSystem2005}
\bibinfo{author}{A.~Andreini}, \bibinfo{author}{C.~Carcasci},
  \bibinfo{author}{S.~Gori}, \bibinfo{author}{M.~Surace},
\newblock \bibinfo{title}{Film {{Cooling System Numerical Design}}:
  {{Adiabatic}} and {{Conjugate Analysis}}},
\newblock in: \bibinfo{booktitle}{Heat {{Transfer}}: {{Volume}} 3},
  \bibinfo{publisher}{ASMEDC}, \bibinfo{address}{San Francisco, California,
  USA}, \bibinfo{year}{2005}, pp. \bibinfo{pages}{9--20}.
  \DOIprefix\doi{10.1115/HT2005-72042}.
\bibitem[{Kirollos and Povey(2015)}]{kirollosEnergyBasedMethodPredicting2015}
\bibinfo{author}{B.~Kirollos}, \bibinfo{author}{T.~Povey},
\newblock \bibinfo{title}{An {{Energy-Based Method}} for {{Predicting}} the
  {{Additive Effect}} of {{Multiple Film Cooling Rows}}},
\newblock \bibinfo{journal}{Journal of Engineering for Gas Turbines and Power}
  \bibinfo{volume}{137} (\bibinfo{year}{2015}) \bibinfo{pages}{122607}.
  \DOIprefix\doi{10.1115/1.4030907}.
\bibitem[{Yang et~al.(2021)Yang, Dai, Rao, and
  Chyu}]{yangMachineLearningApproach2021}
\bibinfo{author}{L.~Yang}, \bibinfo{author}{W.~Dai}, \bibinfo{author}{Y.~Rao},
  \bibinfo{author}{M.~K. Chyu},
\newblock \bibinfo{title}{A machine learning approach to quantify the film
  cooling superposition effect for effusion cooling structures},
\newblock \bibinfo{journal}{International Journal of Thermal Sciences}
  \bibinfo{volume}{162} (\bibinfo{year}{2021}) \bibinfo{pages}{106774}.
  \DOIprefix\doi{10.1016/j.ijthermalsci.2020.106774}.

\end{thebibliography}

\end{document}